\pdfoutput=1
\documentclass[11pt]{article}

\usepackage{ACL2023}

\usepackage{array}
\usepackage[T1]{fontenc}
\usepackage{inconsolata}
\usepackage[utf8]{inputenc}
\usepackage{latexsym}
\usepackage{microtype}
\usepackage{times}

\usepackage[normalem]{ulem}
\useunder{\uline}{\ul}{}
\usepackage{booktabs}
\usepackage{enumitem}
\usepackage{graphicx}
\usepackage{tabularx}
\usepackage{xcolor}
\usepackage{xspace}
\usepackage{colortbl}
\usepackage{caption}
\usepackage{subcaption}
\usepackage{comment}
\usepackage{tablefootnote}
\usepackage{graphicx}
\usepackage{multirow}
\usepackage{tablefootnote}
\newcommand{\dataset}{HRDsAttack\xspace} 

\newif\ifcomments
\commentstrue
\ifcomments
\newcommand{\draftcomment}[3]{{\color{#2}[\textsc{#1} #3]}}
\else
\newcommand{\draftcomment}[3]{}
\fi

\newif\ifdetailed
\ifdetailed
\newcommand{\detailed}[1]{#1}
\newcommand{\abbreviated}[1]{}
\else
\newcommand{\detailed}[1]{}
\newcommand{\abbreviated}[1]{#1}
\fi

\title{
A New Task and Dataset on Detecting Attacks on Human Rights Defenders
}

\author{   
    Shihao Ran \quad    
    Di Lu \\ \bf
    Joel Tetreault \quad
    Aoife Cahill \quad
    Alejandro Jaimes \\
    Dataminr Inc. \quad \\
    \texttt{\{sran,dlu,jtetreault,}\\
    \texttt{acahill,ajaimes\}@dataminr.com} 
}

\begin{document}
\maketitle

\begin{abstract}
The ability to conduct retrospective analyses of attacks on human rights defenders over time and by location is important for humanitarian organizations to better understand historical or ongoing human rights violations and thus better manage the global impact of such events. We hypothesize that NLP can support such efforts by quickly processing large collections of news articles to detect and summarize the characteristics of attacks on human rights defenders. To that end, we propose a new dataset for detecting \textbf{Attack}s on \textbf{H}uman \textbf{R}ights \textbf{D}efender\textbf{s} (\dataset) consisting of crowdsourced annotations on 500 online news articles. The annotations include fine-grained information about the type and location of the attacks, as well as information about the victim(s). We demonstrate the usefulness of the dataset by using it to train and evaluate baseline models on several sub-tasks to predict the annotated characteristics. 

\end{abstract}

\section{Introduction}

\begin{figure}[ht]
    \centering
    \includegraphics[width=1.0\linewidth]{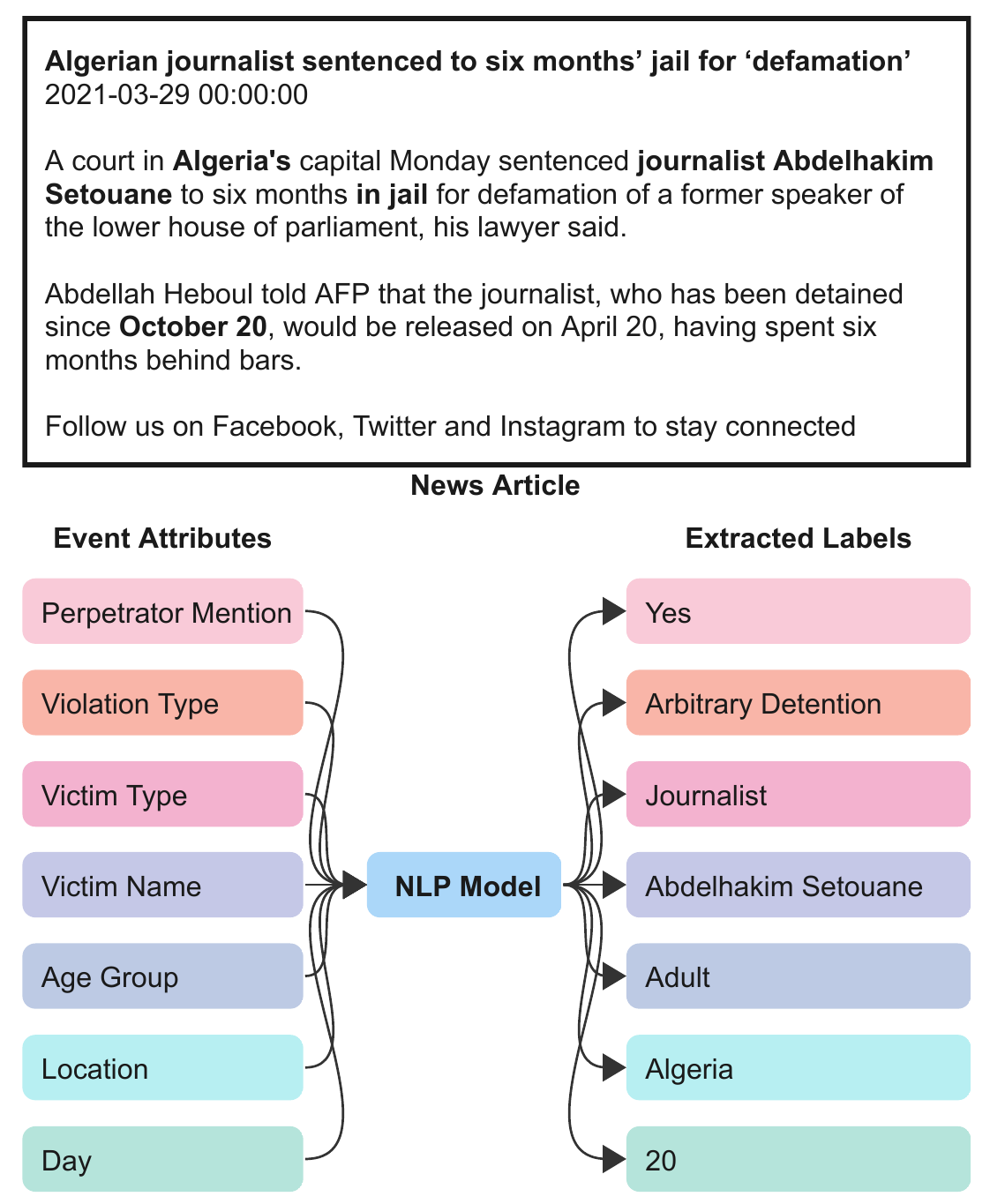}
    \caption{An example of the input/output to an NLP model for extracting event attributes about an attacking event on human rights defenders.}
    \label{fig:ee_hrd}
\end{figure}

It is essential for human rights organizations to track, analyze and summarize attacks on human rights defenders over time and across locations for better personnel protection and situational analysis. To do so, multiple event attributes denoting different aspects of the attacking event need to be extracted from textual sources.  However, this would be a time-consuming process if done manually.
%To do so manually would be a time-consuming, expensive process. 
Figure~\ref{fig:ee_hrd} gives an example of the kinds of information that such organizations need to extract. 
%extracting details about attacks on human rights defenders from news articles.
%One example of such mission-critical applications is identifying event details from online news articles about human rights defenders, as illustrated in Figure~\ref{fig:ee_hrd}. 

%both breadth and depth associated with the critical information around the target event, with which breadth being the number of extracted event attributes and depth being the number of instances or the levels of granularity for each event attribute. For example, the number of victims or different levels of geographical information associated with an attack event. \joel{this sentence starting with "to fully support" needs to be rewritten.  It's like a massive run-on sentence.  Pro tip: read the sentence aloud and see how it flows.  }\shihao{Rewrote.}

% \aoife{I think the Date needs to be changed to Day in Figure \ref{fig:ee_hrd}}

In order to train and evaluate an NLP model to extract this information automatically, a relevant dataset is necessary. The ideal dataset requires accurate annotations for both the breadth (the number of extracted event attributes) and depth (the levels of granularity for each event attribute) of the events. However, all 
%\aoife{can we say all?} \shihao{I think it's safe to say all, added a few other datasets} 
existing Event Extraction (EE) datasets (e.g. ACE05~\cite{doddington2004automatic}, ERE~\cite{song2015light}, ACE05-E~\cite{wadden2019entity}, ACE05-E+~\cite{lin2020joint}) do not contain annotations at a sufficiently fine-grained level. 
%However, many existing Event Extraction (EE) methods~\cite{wadden2019entity,du2020event,lu2021text2event,honnibal2020spacy} trained on existing EE datasets such as ACE05~\cite{doddington2004automatic}, ACE05-E~\cite{wadden2019entity}, ACE05-E+~\cite{lin2020joint}, and CNN/DailyMail~\cite{hermann2015teaching} are not equipped to predict the fine-grained event attributes  shown in Figure~\ref{fig:ee_hrd}. 
Although some existing ontologies and datasets do include annotations related to %cover some relevant aspects of
%\joel{I'm not sure what is meant by "entail relevant information"}\shihao{Rewrote} 
attacking events, e.g. the \textsc{Attack} event type in the ACE05 dataset along with the associated \textsc{Agent} attribute,
they are incomplete with respect to many of the details of interest to human rights organizations and do not contain annotations relevant to victim characteristics or the time/location of the attacking event. As a result, existing open-source EE models trained on these datasets ~\cite{honnibal2020spacy, wadden2019entity, he2019multi} are unable to predict the complete set of relevant information. 

%to precisely pin down an attack event targeting human rights defenders, other event attributes are necessary, such as violation types and perpetrator types, which are not covered by existing ontologies. Meanwhile, the critical information presented in the source needs to be captured in a granular way to ensure the comprehensiveness of the subsequently trained NLP applications, such as different attributes for each victim mentioned in the news article, the varying granularity of the geographical and temporal information of the attack event, and more. Existing open-source EE applications~\cite{honnibal2020spacy, wadden2019entity, he2019multi} and their respective training data can hardly meet all the aforementioned breadths and depths regarding event details.

To mitigate the gap in existing resources,
%and to fully support an efficient modern NLP solution for such tasks, 
we present \dataset, a new dataset containing crowdsourced annotations on 500 online news articles (including article title, article body text, and publication time). Each news article is annotated with 13 different event attributes to capture critical information about attacks on human rights defenders, including the type and location of the attacks, as well as information about the victim(s) and the perpetrator. 
%% Joel - I am commenting out the sentence below because it is in the controbutions section
%Further, we demonstrate the usefulness of \dataset by using it to train and evaluate a strong baseline NLP model based on Question-Answering using the T5 model~\cite{raffel2020exploring} as the backbone in a multi-task setting. 
With \dataset, we hope to support more research opportunities for including NLP in applications related to human rights, as well as for broader AI for Social Good (AI4SG) efforts.

%The enriched annotation regarding each attacking event in \dataset helps make up the resource gap in model training while also posing challenges to existing modeling techniques given the large number of event attributes and sub-tasks. 
%\joel{the sentence before doesn't necessarily flow into the sentence after.  You first say that the additional information is good because you cover all of the event attributes, but it's also complicated to modeling techniques since there are a lot of event attributes...does that mean the dataset is bad?  But then the next sentence skips the bad part and says we train a model.  It just feels like there is some entailment sentence that is missing.  What are you really trying to say?} \shihao{Totally, I removed the confusing part.}

%\aoife{This still needs some work. The new task is presented first, but we've mainly talked about the dataset above}
To summarize, our contributions are threefold: 
\begin{enumerate}
    \item We present a new dataset (\dataset) that includes annotations for fine-grained event details on attacks on human rights defenders. By focusing on expanding the breadth and depth of the attacking event relative to existing EE ontologies, we aim to address the limited scope of existing NLP resources.  The complete ontology for our dataset is shown in Table~\ref{tab:ontology};
    \item We propose a new NLP task to extract fine-grained event details on attacks on human rights defenders. % around Human Rights .\joel{is it fair to include EE in this comparison?} \shihao{Is this better?}
    %\item We present a new dataset \dataset to support model training and evaluation under the proposed task. The complete ontology for our dataset is shown in Table~\ref{tab:ontology}; %\shihao{Removed all the repeated content.}
    \item We demonstrate the usefulness of \dataset with a strong baseline model based on Question Answering (QA) using the T5 model~\cite{raffel2020exploring} as the backbone in a multi-task setting.
\end{enumerate}
The \dataset~dataset along with the code for model training and evaluation is available at \url{https://github.com/dataminr-ai/HRDsAttack}.

\begin{table*}[ht]
\centering
\resizebox{\textwidth}{!}{%
\begin{tabular}{|l|l|l|l|}
\hline
\rowcolor[HTML]{ECF4FF} 
\multicolumn{1}{|c|}{\cellcolor[HTML]{ECF4FF}Category} & \multicolumn{1}{c|}{\cellcolor[HTML]{ECF4FF}Event Attribute} & \multicolumn{1}{c|}{\cellcolor[HTML]{ECF4FF}Labels} & \multicolumn{1}{c|}{\cellcolor[HTML]{ECF4FF}Label Definitions} \\ \hline
\midrule
\multirow{10}{*}{Perpetrator} & \multirow{2}{*}{\begin{tabular}[c]{@{}l@{}}Perpetrator \\ Mention\end{tabular}} & Yes & There is one or more explicit mention of the perpetrator in the news article. \\ \cline{3-4} 
 &  & No & There is no explicit mention of the perpetrator in the news article. \\ \cline{2-4} 
 & \multirow{8}{*}{\begin{tabular}[c]{@{}l@{}}Perpetrator \\ Type\end{tabular}} & State Security Forces & Anyone employed by or representing a state institution. \\ \cline{3-4} 
 &  & Other State Actors & Other actors that are a part of the state or other non-military authorities of a state. \\ \cline{3-4} 
 &  & \begin{tabular}[c]{@{}l@{}}Other non-state\\ actors\end{tabular} & \begin{tabular}[c]{@{}l@{}}Other actors /Private actors that are not a part of the state and act without the state’s \\ permission, support, or acquiescence.\end{tabular} \\ \cline{3-4} 
 &  & \begin{tabular}[c]{@{}l@{}}Other actors with \\ permissions\end{tabular} & \begin{tabular}[c]{@{}l@{}}Armed actors that are not a part of the state but act with the state’s permission, support \\ or acquiescence.\end{tabular} \\ \cline{3-4} 
 &  & \begin{tabular}[c]{@{}l@{}}Other actors without \\ permissions\end{tabular} & Other actors that are not a part of the state. \\ \cline{3-4} 
 &  & Regional Organizations & Person or group working for a regional or international organization. \\ \cline{3-4} 
 &  & Insufficient Information & \begin{tabular}[c]{@{}l@{}}There is insufficient information available to determine one of the categories \\ described above.\end{tabular} \\ \cline{3-4} 
 &  & None & Not applicable, when Perpetrator mention is No. \\ \hline
\multirow{7}{*}{Violation} & \multirow{7}{*}{Violation Type} & Arbitrary Detention & Arrest or detention not in accordance with national laws. \\ \cline{3-4} 
 &  & Enforced Disappearance & \begin{tabular}[c]{@{}l@{}}Unlawful deprivation of liberty enforced or authorized by the state, that is\\ not acknowledged by the state or the location of the victim is kept secret.\end{tabular} \\ \cline{3-4} 
 &  & Killing & Unlawful death inflicted upon a person with the intent to cause death or serious injury. \\ \cline{3-4} 
 &  & Kidnapping & Deprivation of liberty that is not enforced or authorized by the state. \\ \cline{3-4} 
 &  & Torture & \begin{tabular}[c]{@{}l@{}}The action or practice of inflicting severe pain or suffering on someone as a \\ punishment or in order to force them to do or say something.\end{tabular} \\ \cline{3-4} 
 &  & Other & \begin{tabular}[c]{@{}l@{}}Sexual violence or other acts causing or intending to cause harm, such as coercion or\\ discrimination.\end{tabular} \\ \cline{3-4} 
 &  & Unknown & \begin{tabular}[c]{@{}l@{}}No harmful acts were conducted or there is insufficient information to\\ determine the harmful acts.\end{tabular} \\ \hline
\multirow{15}{*}{Victim} & Victim  Name & - & Name of the victim. \\ \cline{2-4} 
 & \multirow{4}{*}{Victim Type} & Human Rights Defender & \begin{tabular}[c]{@{}l@{}}A person exercising their right, to promote and strive for the protection and \\ realization of human rights and fundamental freedoms.\end{tabular} \\ \cline{3-4} 
 &  & Trade Unionist & \begin{tabular}[c]{@{}l@{}}A person exercising their right to form and join trade unions to protect\\ their interests.\end{tabular} \\ \cline{3-4} 
 &  & Journalist & \begin{tabular}[c]{@{}l@{}}A person observing events, statements, policies, etc. that can affect society, with the purpose\\ of systematizing such information to inform society.\end{tabular} \\ \cline{3-4} 
 &  & Insufficient Information & \begin{tabular}[c]{@{}l@{}}There is insufficient information available to make select one of the categories \\ described above.\end{tabular} \\ \cline{2-4} 
 & \multirow{2}{*}{\begin{tabular}[c]{@{}l@{}}Victim Population \\ Type\end{tabular}} & Individual & A named individual victim. \\ \cline{3-4} 
 &  & Multiple & Multiple unnamed individuals. \\ \cline{2-4} 
 & \multirow{4}{*}{Victim Age Group} & Adult & Age \textgreater{}= 18. \\ \cline{3-4} 
 &  & Child & Age \textless 17. \\ \cline{3-4} 
 &  & Other & A mixture of age groups, when Victim Population Type is Multiple. \\ \cline{3-4} 
 &  & Unknown & There is insufficient information available to determine the age group. \\ \cline{2-4} 
 & \multirow{4}{*}{Victim Sex Group} & Man & Male. \\ \cline{3-4} 
 &  & Woman & Female. \\ \cline{3-4} 
 &  & Other & Other gender types. \\ \cline{3-4} 
 &  & Unknown & There is insufficient information available to determine the sex group. \\ \hline
\multirow{3}{*}{Location} 
 & Country & - & Country in which the attack occurred \\ \cline{2-4} 
 & Region & - & Region in which the attack occurred, such as a state or a province \\ \cline{2-4} 
 & City & - &  City in which the attack occurred\\ \hline
\multirow{3}{*}{Time} & Year & - &  Year the attacking event occurred \\ \cline{2-4} 
 & Month & January, ..., December & Month the attacking event occurred \\ \cline{2-4} 
 & Day & 1, 2, 3, ..., 31 & Day (of the month) the attacking event occurred \\ \hline
\end{tabular}%
}
\caption{Labeling ontology of \dataset.}
\label{tab:ontology}
\end{table*}
\section{Related Work}
% Our work relates to Event Extraction, NLP in Human Rights, and Question Answering.
%\aoife{here we talk about EE as being the related work, but really the related work are the existing datasets}
\subsection{Event Extraction}
Event Extraction (EE) is an NLP task that aims to extract key information such as \textit{who, what, where, and when} from a text. The most commonly used dataset for EE is the ACE05 English corpus~\cite{doddington2004automatic} which consists of 33 event types and 22 event argument roles across 599 documents from newswires, web blogs, and broadcast conversations. While the ACE ontology covers a large range of event types, only two of them are related to attacking events: the \textsc{Life.Injure} event and the \textsc{Conflict.Attack} event. Some of the other datasets that focus on extracting event triggers or event arguments are based on the ACE05 ontology%\joel{are all EE datasets based on the ACE05 ontology?  If so maybe we just leave that phrase out} \shihao{No, there are other datasets based on other ontologies that are expansions of ACE}
~\cite{wadden2019entity,lin2020joint}, and only cover limited aspects of the information that \dataset covers, e.g. the \textsc{Attacker} and \textsc{Target} attributes in the \textsc{Life.Injure} and \textsc{Conflict.Attack} events. The Armed Conflict Location and Event Data (ACLED) dataset ~\cite{10.2307/20798933} covers political violence and protest events with annotations for event type, actors and targets, but it does not cover victim-dependent attributes.
In comparison, 
% for an accurate understanding of the event and for supporting a comprehensive event-detail extraction task, 
\dataset focuses on attacking events on human rights defenders and provides more event attributes for the attacks, along with more granular information regarding each event attribute.
% The Armed Conflict Location and Event Data (ACLED) dataset ~\cite{10.2307/20798933} is the most similar one to \dataset, it covers political violence and protest events with annotations for event type, actors and targets. 

In terms of modeling approaches, early work on EE  formulated the task as a token-based classification problem which leveraged different types of features~\cite{ahn2006stages,liao2010filtered,liao2010using,li2013joint}. More recent approaches focus on applying neural models to EE tasks, such as CNNs~\cite{chen2015event}, RNNs~\cite{liu2019exploiting}, and other advanced model structures~\cite{nguyen2019one,zhang2019extracting}.

% \joel{Going back to my comments in Section 1 - we get thrugh section 2 still not knowing what is our ontology.  We keep hearing how it is different from prior ones but the reader still hasn't seen what it is.  This should be done earlier in my opinion.  Then the comparisons make way more sense.}

% \joel{Related work should have subsections.  You have three contributions so in my head you should note if there are similar tasks, similar datasets, and similar models, and anything similar in the AI4SG space.}

\subsection{NLP Research for Human Rights}
Existing NLP research resources around event detection and extraction related to Human Rights are extremely limited. Previous work has focused on identifying potential human rights abuse incidents from social media posts \cite{alhelbawy2020nlp,pilankar2022detecting}, alongside more general applications such as detecting abusive language~\cite{golbeck2017large,djuric2015hate,aroyo2019crowdsourcing}, or procedure-focused applications (e.g. data modeling processes for human rights data~\cite{miller2013digging,fariss2015human}), or predicting judicial decisions of the European Court of Human Rights using NLP~\cite{o2019predicting}.  To our knowledge, there are no event extraction datasets which target human rights issues, which makes \dataset a first in this research area.  %\joel{what do you think of this last sentence?  Trying to keep things to th e point and easy to understand} \shihao{Looks good!}

%Many existing resources for supporting EE tasks are not dedicated to the Human Rights domain, with limited relevant sample size and event attribute coverage, such as the aforementioned ACE datasets. In comparison, our work focuses on building a quality dataset that better supports the automated detection of detailed characteristics of attacks on human rights defenders by focusing on the target domain, with more annotated samples and more detailed event attributes.

%\subsection{Question Answering}
% Due to recent advances in large Pre-trained Language Models (PLMs) such as BERT \cite{devlin2018bert}, many NLP applications have been recently framed as Question-Answering (QA) tasks \cite{chali2011improving,dwivedi2013research,ansari2016intelligent,lende2016question}, including EE~\cite{liu2020event,du2020event,zhang2020question}. QA-based methods leverage the inherited knowledge in large PLMs to further assist the inference process in a traditional classification setting \cite{zhang2020question,du2020event}. In \citet{liu2020event}'s work, the EE task is modeled as a Machine Reading Comprehension (MRC) task and uses unsupervised pre-training to automatically generate topic-relevant and context-dependent natural questions via a style-transfer process. In our work, we adopt a Sequence-to-Sequence learning framework and use T5 as the backbone model.
\section{Dataset}
In this section, we describe the construction of the \dataset dataset, which contains 500 annotated news articles, including article title, article body text, and publication time. We select news articles as the data source rather than other data sources (such as social media posts) since online news articles generally have higher accessibility, better trustworthiness of the source, and longer content length.

% the following reasons:
% \begin{itemize}
%     \item News articles are generally more accessible globally in comparison to social media. Some social media platforms are only popular or available in certain regions, but news articles are distributed through local news outlets globally.
%     \item News articles generally contain more information than social media posts which often have a character limit.
%     \item News articles are often written by trusted news agencies or reporters, hence the trustworthiness of the information is higher than the chatters on social media or other data sources.
% \end{itemize}

% \joel{note if space is an issue we can shrink this preamable about "why news" considerably.  Really since it's a retrospective, one would want something reliable and real-time isn't going to be a }
In our work, we sample online news articles from the GDELT database\footnote{\url{https://www.gdeltproject.org/}}, which we discuss in more detail in Section \ref{sec:data_sampling}.
% We first describe the annotation labels in the dataset (Section~\ref{sec:annotation_labels}), followed by the data sampling strategy (Section~\ref{sec:data_sampling}) for building \dataset. We then summarize the statistics of the dataset in Section~\ref{sec:data_statistics}.

\subsection{Annotation Labels}
\label{sec:annotation_labels}
To ensure the comprehensiveness of the annotations regarding capturing event details, we first identify the event attributes or labels required for annotation. As shown in Table \ref{tab:ontology}, according to the UN Human Rights SDG 16.10.1 Guidance Note\footnote{\url{https://www.ohchr.org/Documents/Issues/HRIndicators/SDG_Indicator_16_10_1_Guidance_Note.pdf}}, we identify the following 5 categories of attributes: \textsc{Perpetrator}, \textsc{Violation}, \textsc{Victim}, \textsc{Location}, and \textsc{Time}. Each category has one or more associated event attributes, all denoting key information about the primary event described in the original article\footnote{All label values for each event attribute are prescribed by the SDG 16.10.1 Guidance Note.}. If there are multiple events mentioned in the article, only the primary event (i.e. the event that happened closest to the publication time) is annotated. We also specify that the \textsc{Victim} category could have multiple entries per article, while other categories can only have one entry per article (i.e. only one entry for the primary attack event). The ontology for the annotation labels is shown in Table~\ref{tab:ontology}.

\subsection{Data Sampling}
\label{sec:data_sampling}
To build \dataset, we first scrape 80,112 online news articles in the time range of 2019/09/01 to 2022/05/01 from the GDELT database following the CAMEO codebook~\cite{schrodt2012conflict}, a standard framework for coding event data~\cite{yuan2016modeling}. %\joel{what is the CAMEO codebook and why is it important to note?} \shihao{Added}
These scraped news articles are identified as relevant to human rights defenders by an existing human rights monitoring workflow. 
% \shihao{UN OHCHR de-anonymized.}

% a human rights organization.\footnote{Name of organization withheld for anonymization purposes.} 

% From our pilot annotation tasks, we identified imbalanced distributions for some of the event attributes, such as victim type and violation type. To address this issue, we apply keyword matching to the entire GDELT dataset and locate the samples that have a higher probability of containing the minority event attributes (such as \textsc{Trade Unionist} under \textsc{Victim Type}). We then apply stratified sampling across these sub-populations instead of random sampling across the entire scraped sample set. By doing so, the sampled dataset has a more balanced distribution across all class labels. 

During our pilot studies, we identified a data imbalance issue from the annotations under random sampling. Specifically, we observed significantly skewed label distributions in event attributes \textsc{Violation Type} and \textsc{Victim Type}, the minority classes being \textsc{Torture} and \textsc{Kidnapping} for \textsc{Violation Type}, and \textsc{Human Rights Defenders} and \textsc{Trade Unionists} for \textsc{Victim Type}. To address this issue, we apply keyword filtering
%\aoife{Is this keyword filtering different from the keyword filtering that's described later to increase the sample of the minority classes?} \shihao{It's the same one. Shifted this part}
 and targeted sampling to ensure \dataset is well-balanced across classes in each event attribute. 

To include more samples with a higher probability of containing events associated with these minority attributes, we first reduce the original 80,112 samples into four smaller, targeted sample sets.
%\joel{what are sub-groups?  what does it mean to augment?} \shihao{Rewrote to make it easier to understand} 
Each targeted sample set corresponds to the articles that contain the keyword for each of the minority classes. We then randomly sample 25 articles from each targeted sample set to form a batch of 100 samples for each round of full annotation. Table~\ref{tab:keywords} shows the keywords used for minority class targeted sampling.

\begin{table}[h]
\centering
\begin{tabular}{cc}
\hline
\rowcolor[HTML]{ECF4FF}
\cellcolor[HTML]{ECF4FF}Minority Event Attribute     & \cellcolor[HTML]{ECF4FF}Keyword \\ \hline
\midrule
Torture              & torture          \\ \hline
Kidnapping         & kidnapping          \\ \hline
Human Rights Defenders                       & human right             \\ \hline
Trade Unionists                  & trade union             \\ \hline
\end{tabular}
\caption{Keyword for each minority class used in keyword filtering and targeted sampling.}
\label{tab:keywords}
\end{table}

\subsection{Annotation Process}

The annotation is done by qualified workers (Turkers) on Amazon Mechanical Turk (AMT). We design and implement a separate qualification task to recruit top-performing Turkers, and we only release the full annotation tasks to the Turkers that surpass a predefined performance bar based on the qualification tasks. 

\subsubsection{Qualification Tasks}
For the qualification task, all US-based Turkers that have a HIT (Human Intelligence Task \footnote{A HIT represents a single, self-contained, virtual task that a Turker can work on, submit an answer, and collect a reward for completing.}) approval rate greater than 90\% and a total number of HITs approved greater than 500 are able to participate. 
%\joel{we didn't filter by their acceptance score or anything?} \shihao{Good catch, added} 
In the qualification task, we sample three different news articles and ask all participant Turkers to annotate every event attribute for each news article through three questionnaires (each HIT contains three questionnaires, one for each news article). We then evaluate their performance on this annotation task. All three news articles are also annotated by domain experts, and we use their annotations as the ground truth answers for calculating the Turker accuracy. We only recruit Turkers who have 75\% or higher average accuracy across all three news articles. We launched three rounds of qualification tasks with 50 assignments in total, and ten Turkers passed the qualification tasks.

The instructions and the task interface for the qualification tasks are shown in Figures~\ref{fig:qua_1} to \ref{fig:qua_8} in Appendix~\ref{appx:annotation_ui}.
\subsubsection{Full tasks}

In the full task, each HIT only contains a single news article. The instructions and the annotation interface are identical to the qualification task. We launched all 500 samples in 5 batches, each batch containing 100 HITs. During our pilot studies, we did not observe a significant quality improvement with replication factor 3 due to relatively high agreement scores between the Turkers (Table~\ref{tab:worker_agreement} in Appendix~\ref{appx:kappa_score}). We hypothesize that this is because the annotation task itself is highly objective. Therefore, we did not apply replication factors during the full task.

We compensate each Turker with \$7.50 per assignment in the qualification task (three news articles per assignment) and \$2.00 per assignment in the full task (one news article per assignment). We also provide an additional bonus to all participant Turkers of %based on the number of assignments they completed with 
 \$0.5 per assignment. The final pay rate is \$15.00 per hour, which is over the US national minimal wage of \$7.50\footnote{\url{https://www.dol.gov/general/topic/wages/minimumwage}}.

The annotation instructions and the task interface for the full tasks are shown in Figures~\ref{fig:full_1} to \ref{fig:full_4} in Appendix~\ref{appx:annotation_ui}.
\subsection{Data Statistics}
\label{sec:data_statistics}
To create a benchmark dataset from \dataset, we randomly split the 500 annotated samples into train, dev, and test set with a 3:1:1 ratio. Table~\ref{tab:data_stats} shows the statistics of the splits.  A breakdown of the label-level statistics for each event attribute can be found in Table~\ref{tab:label_stats} in Appendix~\ref{appx:label_stats}.

% \begin{figure}[!ht]
%     \centering
%     \includegraphics[width=\linewidth]{figures/table_splits_stats.png}
%     \caption{\shihao{To replace} Statistics of \dataset dataset.}
%     \label{tab:data_stats}
% \end{figure}

\begin{table}[h]
\centering
\resizebox{\columnwidth}{!}{%
\begin{tabular}{lrrrr}
\hline
\rowcolor[HTML]{ECF4FF}
 & \multicolumn{1}{c}{\cellcolor[HTML]{ECF4FF}Train} & \multicolumn{1}{c}{\cellcolor[HTML]{ECF4FF}Dev} & \multicolumn{1}{c}{\cellcolor[HTML]{ECF4FF}Test} & \multicolumn{1}{c}{\cellcolor[HTML]{ECF4FF}Total} \\ \hline
 \midrule
No. of Articles        & 300     & 100    & 100      & 500      \\ \hline
Total No. of Tokens    & 287,911 & 97,038 & 124,658  & 509,607  \\ \hline
Avg. No. of Tokens  & 959.70  & 970.38 & 1,246.58 & 1,019.21 \\ \hline
Total No. of Victims   & 687     & 272    & 204      & 1,163    \\ \hline
Avg. No. of Victims & 2.29    & 2.72   & 2.04     & 2.33     \\ \hline
\end{tabular}%
}
\caption{
Textual statistics of \dataset splits. The average number of tokens and victims is averaged per news article. 
}
\label{tab:data_stats}
\end{table}
\section{Our Model}
With the construction of \dataset, we now turn to developing a model for the task.  We noted earlier that existing state-of-the-art EE models are not suitable as baselines, as they rely on extensive human annotations based on token-level annotations, hence cannot easily be re-trained and evaluated on this dataset. For instance, AMR-IE~\cite{zhang2021abstract} and GraphIE~\cite{qian2018graphie} are trained on the ACE05 dataset and ERE dataset. Some recent research casts the EE task as QA tasks or Seq2seq tasks, such as RCEE\_ER~\cite{liu2020event} and Text2Event~\cite{lu2021text2event}. In this section, we propose a new model for extracting fine-grained details regarding attacks on human rights defenders.

\subsection{Overall Framework}
\label{sec:overall_framework}

\begin{figure*}
    \centering
    \includegraphics[width=1.0\linewidth]{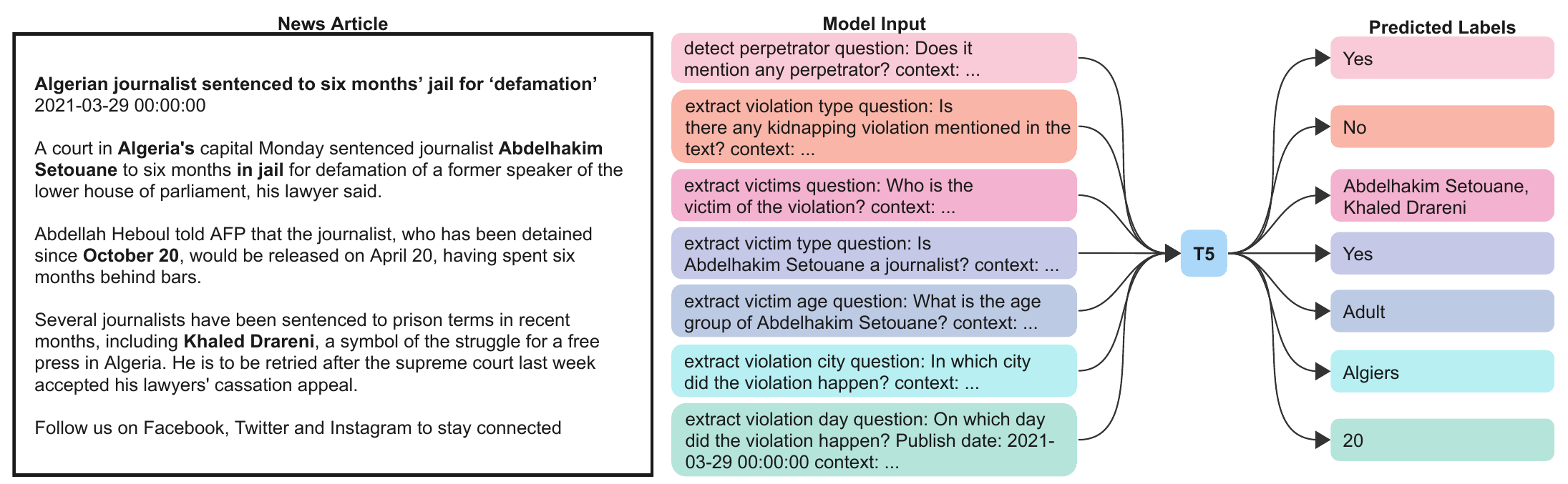}
    \caption{Overall framework of the proposed Sequence-to-Sequence Question-Answering model.}
    \label{fig:overall_framework}
\end{figure*}

%based on a 

Given the limited amount of training data and the range and variety of event attributes, we propose using a single Seq2Seq Question Answering (QA) model.
Training a unified model has the advantageous property that it shares the training data across all the sub-tasks thus potentially leading to better  performance for each sub-task. Figure~\ref{fig:overall_framework} shows the overall framework of our proposed baseline model.
%\joel{nowhere in this paper do we say we are looking to build a model specific for this, just to make some baselines.  I think that should be clear.  Given the amount of space dedicated to this model, it seems like the former.  That's ok, but we should be clear about it upfront.  I guess in prior resources I would state all the different baselines I would run and then have at the end the model that is specific to the task. } \shihao{Added the keyword 'baseline'}

We formulate all of the subtasks as a generation task following T5~\cite{raffel2020exploring}, which proposes reframing all NLP tasks into a unified text-to-text format. The input to the T5 model is a natural language sentence composed of (1) a task prefix (e.g. \textit{`extract victims'}), (2) an attribute-oriented question (e.g. \textit{`Who is the victim of the violation?'}), and (3) a context which is the original article. 
The output is a text string which explicitly refers to the value of the concerned event attribute (e.g. \textit{`Abdelhakim Setouane'}).
%\aoife{Should we maybe only include the example that appears in the text from the introduction?}
% The output is the answer to the input question. 
% For example, in order to extract the victim names from the article, we concatenate the task prefix \textit{extract victims}, the question \textit{Who is the victim of the violation?}, and the original article as the input context, and the model generates \textit{Abdelhakim Setouane, Khaled Drareni} as the output.

\subsection{Input-Output Design}
\label{sec:question}
We group the event attributes into three categories: general article-dependent attributes, victim-dependent attributes, and publication time-dependent attributes, and we design input and output formats for them respectively. For all of the three categories, the output is a text string that explicitly refers to the value of the relevant event attribute, e.g. \textit{`Yes'} for \textsc{Perpetrator Mention}, or \textit{`state security forces'} for \textsc{Perpetrator Type}. The input formats for the three categories have minor differences~\footnote{The complete lists of input and output formats are provided in Table~\ref{tab:general_article_dependent_classes} in Appendix~\ref{appx:in_output}.}:

% \joel{it seems that a lot of text is being devoted to something that could be placec in a table for 4.2.  There is a lot of repetition and stating of the obvious.  Can we make a table to represent this information?  Could even be a table per section.  I think the main thing is that this section could be shrunk, or even put in the appendix since it's not the main focus of the paper - the dataset is.}

\begin{itemize}
\item \textbf{General Article-dependent Attributes:}
Most of the event attributes depend on the general information
%\joel{what is general information?} \shihao{Explained}
contained within the article (i.e. do not rely on additional input other than article's body text). These include \textsc{Perpetrator Mention}, \textsc{Perpetrator Type}, and \textsc{Violation Type}. For these attributes, the input is the concatenation of a task prefix, an attribute-oriented question, and the original article (e.g. the top three examples in Figure~\ref{fig:overall_framework}).
\item \textbf{Victim-dependent Attributes:}
Some event attributes, such as \textsc{Victim Sex Type}, depend on the information related to a specific victim. Thus we incorporate the victim name into the input question, as exemplified in the fourth and fifth examples in Figure~\ref{fig:overall_framework}.

% In these cases, the question for each class is customized for each victim by adding the victim's name to the question template\footnote{We use the gold-labeled victims for training and evaluation.}. For example, to predict the age group of a specific victim, we ask the model the question \textit{‘What is the age group of \{VICTIM\_NAME\}?’} with \textit{\{VICTIM\_NAME\}} replaced by a specific victim name. The answer is one of the predefined labels: \{\textit{adult, child, other, unknown}\}.
\item \textbf{Publication Time-dependent Attributes:}
%$\aoife{I don't think this is quite right? The Year, Month and Day attributes refer to the date of the attack event, not the publication date, right? There is a relationship to the publication date, but I don't know that it's a direct dependency? I think in Figure 2, the Predicted label for the date should be October 20 (not just 20?) based on what's bolded in the original article text?}\di{rewrote the paraphrase. those `date` are supposed to be `day`, we've corrected some of them in the paper, but forgot the one in the figure.}
In some cases, the \textsc{Year}, \textsc{Month}, and \textsc{Day} attributes related to the attack event are not explicitly present in the article, and we need to infer them based on a combination of the article publication time and the relevant time mentioned in the article (e.g. \textit{last month, two weeks ago, yesterday}). 
The article publication time is available as metadata in the GDELT dataset (e.g. \textit{2021-03-29 00:00:00}). For these attributes, we add publication time information into the input, as shown in the last example of Figure~\ref{fig:overall_framework}.
% Note that the time when the attack event happened might not necessarily be the same or around the time when the article was published. And an explicit mention of the exact time (year, month, and day) might not be available in the article, either. 
% In some cases, the article only mentions a relevant time of the event, such as \textit{last month, two weeks ago, yesterday}. To get the accurate absolute time of the event, we need to infer the year, month, and day of the event based on the article publication time combined with the relevant time mentioned in the article. In these cases, the question for each class is appended with the date when the article was published. The date information is available as metadata in the GDELT dataset and is in the format like ‘2021-06-20 00:00:00’. 
% Table ~\ref{tab:publication_time_dependent_classes} shows the summary of the predefined questions and answers for classes dependent on publication information:
% . But during real-time inference, the model first extracts victim names, and then uses the predicted victim names to generate the questions for victim-dependent classes.}. 

\end{itemize}

\noindent\textbf{Task Prefix.}
\label{sec:prefix}
Following the multi-task setting in the original T5 work, we add a task prefix at the beginning of the input text. The task prefix is used to instruct the T5 model to perform a particular task. It could be any arbitrary text. In our work, we use a brief task description as the task prefix for each event attribute, e.g. \textit{`detect perpetrator'} for \textsc{Perpetrator Mention} or \textit{`extract violation type'} for \textsc{Violation Type} (Figure~\ref{fig:overall_framework}). The complete list of all the task prefixes is shown in Table~\ref{tab:task_prefix} in Appendix~\ref{appx:in_output}.

\subsection{Long Document Resolution}
\label{sec:long}

The maximum input length allowed by the T5 model is 512 tokens, but around 75\% of the articles from the GDELT dataset exceed that length limit. We explore two options to deal with articles with more than 512 tokens: \textbf{Truncation} and \textbf{Knowledge Fusion}. Additional methods for handling long documents are discussed in Appendix \ref{appx:long_doc}. 
% \shihao{Related methods added for Reviewer question 3B.}

\begin{figure}
    \centering
    \includegraphics[width=1.0\linewidth]{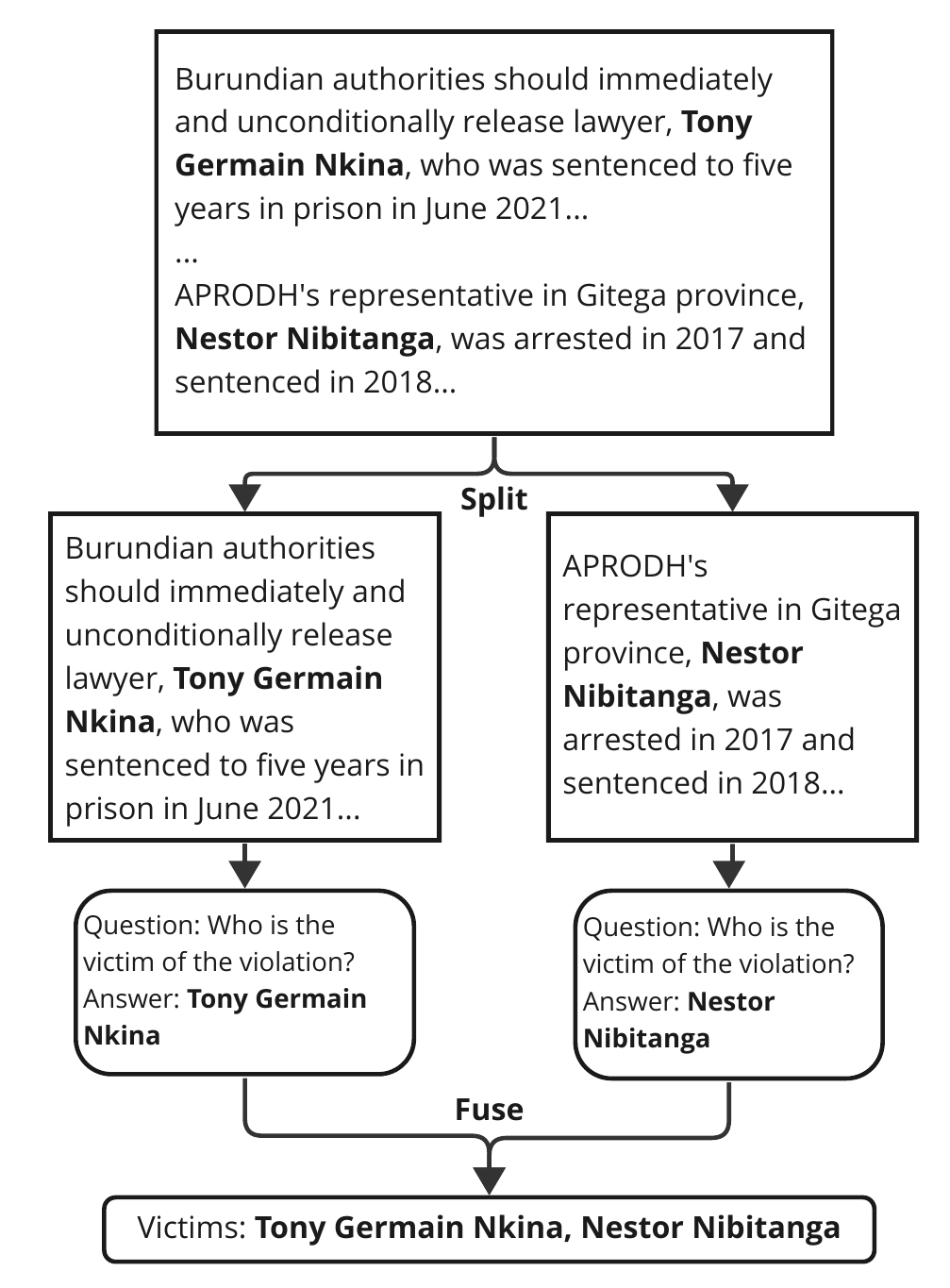}
    \caption{Knowledge Fusion approach.}
    \label{fig:knowledge_fusion}
\end{figure}

\begin{description}
\item \textbf{Truncation.} We only use the first 512 tokens of the input text. The articles from GDELT are news articles, and the first several sentences from a news article usually contain the most important information. Thus a simple solution is to truncate the article and ignore the cut content.
\item \textbf{Knowledge Fusion.} To mitigate the information loss in the Truncation method, we adopt a split-fuse approach~(Figure~\ref{fig:knowledge_fusion}) by (1) splitting the documents into short paragraphs using the spaCy~\cite{honnibal2020spacy} tokenizer\footnote{We use the en\_core\_web\_sm spaCy pipeline.}; (2) applying the model to each of the paragraphs; and then (3) merging the results from each paragraph to obtain the final results for the original article. For event attributes that allow more than one value (e.g. \textsc{Victim Names}), we keep all of the unique results, and for other attributes, we only keep the one with the highest confidence score (beam search score).
\end{description}

\section{Experiments}
% In this section, we describe the experiments we conducted with \dataset and the evaluation results from the strong baseline models.

\subsection{Evaluation Metrics}
We consider the following metrics for evaluating different event attributes:
\begin{itemize}
    \item \textbf{Precision, Recall, and F1 Score}: we use Precision, Recall, and F1 score to evaluate the model performance on \textsc{Perpetrator Mention} and \textsc{Violation Type}.
    \item \textbf{Accuracy}: we use accuracy (i.e. percentage correct) to evaluate the model performance on \textsc{Perpetrator Type}, \textsc{Victim Type}, \textsc{Victim Sex Type}, \textsc{Victim Age Group}, \textsc{Country}, \textsc{Region}, \textsc{City}, \textsc{Year}, \textsc{Month}, and \textsc{Date}.
    \item \textbf{Fuzzy Match Precision, Recall, and F1 Score}: For the \textsc{Victim Name} attribute, we use precision, recall, and F1 score based on exact matching and fuzzy matching, respectively. For exact matching, one predicted victim name is counted as correct only if it exactly matches with a victim name in the ground truth. For fuzzy matching, one predicted victim name is counted as correct if it has overlapping tokens with a victim name in the ground truth. For example, a predicted victim name \textit{Jordan} is counted as correct when it matches with a ground truth name \textit{Michael Jordan}. 
    % It is also counted as correct if it contains a victim name in the ground truth.
\end{itemize}

\subsection{Baseline Models}
We consider the following models in our evaluation:
% \joel{I wouldn't say off-the-shelf, I would just use DyGIE}
% \joel{Also, can we retrain pr fine-tune DyGIE on this dataset?  It's the most natural thing to do }
% \di{the model was trained on token-level annotations, and not designed for those victim-attributes extraction}
\begin{itemize}
    \item \textbf{DyGIE++}~\cite{wadden2019entity}: a joint Information Extraction (IE) model and we use the checkpoint trained on the ACE05 dataset. 
    It requires mapping from the ACE event ontology\footnote{The ACE ontology covers event types such as \textsc{Attack} and \textsc{Injure}.} to \dataset. As a result, it only covers two attributes: \textsc{Perpetrator Mention} and \textsc{Victim Name} as there is no available mapping for the other event attributes in \dataset.
    \item \textbf{T5 w/ Truncation}: our proposed T5-based model with truncation. 
    % Only the first 512 tokens are fed into the T5 model.
    \item \textbf{T5 w/ Knowledge Fusion}: our proposed T5-based model with knowledge fusion. 
    % The original article is first split into paragraphs, and each paragraph is fed into the T5 model sequentially. The final predictions for the article are based on the fused results of the predictions in each paragraph.
    \item \textbf{Hybrid (final model)}: a hybrid model based on T5 w/ Truncation and T5 w/ Knowledge Fusion. The model only applies knowledge fusion to \textsc{Perpetrator Mention}, \textsc{Victim Name}, and \textsc{Victim Age Group} attributes. This hybrid strategy is decided based on the evaluation results on the dev set.
\end{itemize}

We recognize that it would be ideal to have more baseline models for comparison, such as a retrained version of DyGIE++ on \dataset. However, many existing EE models are trained on token-level annotations and are not designed for the additional event attributes that \dataset covers (e.g. \textsc{Victim Types}). Therefore, we had to design a specialized model for this task.  We plan to benchmark more Sequence-to-Sequence based models on \dataset in future work.
%\joel{I'm concerned that we don't have more baselines.  Could we have retrained any EE model on our dataset?  }
%
%\joel{I saw the comment (by Di) on why DyGIE can't be retrained.  I think you should put that explanation in the paper because someone would rightfully ask.}\shihao{added.}

\subsection{Training Implementation}
We use the T5-large checkpoint~\footnote{\url{https://huggingface.co/t5-large}} provided by Huggingface~\cite{mromero2021t5-base-finetuned-question-generation-ap} to initialize the model and all experiments are run on a single AWS g5.xlarge instance. The AWS g5.xlarge instance is equipped with a single NVIDIA A10G GPU with 24 GB of GPU memory. Table~\ref{tab:training_hp} shows the hyperparameters we use to train the model.
 
\begin{table}[h]
\centering
\begin{tabular}{cc}
\hline
\rowcolor[HTML]{ECF4FF}
\cellcolor[HTML]{ECF4FF}Hyperparameter     & \cellcolor[HTML]{ECF4FF}Value \\ \hline
\midrule
Learning rate               & 1e-4           \\ \hline
Learning rate decay         & 1e-5           \\ \hline
Epoch                       & 20             \\ \hline
Batch size                  & 4              \\ \hline
Gradient accumulation steps & 16             \\ \hline
\end{tabular}
\caption{Hyperparameter settings for model training.}
\label{tab:training_hp}
\end{table}

\begin{table*}[!ht]
\centering
\resizebox{\textwidth}{!}{%
\begin{tabular}{llcccc}
\hline
\rowcolor[HTML]{ECF4FF}
\cellcolor[HTML]{ECF4FF}Event Attribute                                          & \cellcolor[HTML]{ECF4FF}Metric                & \multicolumn{1}{l}{\cellcolor[HTML]{ECF4FF}DyGIE++} & \cellcolor[HTML]{ECF4FF}T5 w/ Truncation & \cellcolor[HTML]{ECF4FF}T5 w/ \cellcolor[HTML]{ECF4FF}Knowledge Fusion & Hybrid \\ \hline
\midrule
\multicolumn{1}{c}{\multirow{3}{*}{Perpetrator Mention}} & Precision &   \textbf{100.00} &  93.68  &             93.81           &   93.81     \\
\multicolumn{1}{c}{} & Recall                & 36.54  & 97.80 & \textbf{100.00} & 100.00 \\
\multicolumn{1}{c}{} & F1                    &  53.52 & 95.70  & \textbf{96.81} & 96.81 \\ \hline
Perpetrator Type     & Accuracy              &  - & \textbf{62.00} & 60.00 & 62.00 \\ \hline
\multirow{6}{*}{Victim Name}                             & Exact Match Precision &                9.41                   &           \textbf{75.61}       &              59.30          &     59.30   \\
                     & Exact Match Recall    & 9.19  & 24.03 & \textbf{39.53} & 39.53 \\
                     & Exact Match F1        &  9.30 & 36.47 & \textbf{47.44} & 47.44 \\
                     & Fuzzy Match Precision & 17.65  & \textbf{85.37} & 63.95 & 63.95 \\
                     & Fuzzy Match Recall    & 17.24  & 27.13 & \textbf{42.64} & 42.64 \\
                     & Fuzzy Match F1        &  17.44 & 41.18 & \textbf{51.16} & 51.16 \\ \hline
Victim Type          & Accuracy              & - & \textbf{72.41} & 71.67 & 72.41 \\ \hline
Victim Sex Type      & Accuracy              & - & \textbf{89.66} & 86.67 & 89.66 \\ \hline
Victim Age Group     & Accuracy              & - & \textbf{93.10} & 92.50 & 92.50 \\ \hline
\multirow{3}{*}{Violation Type}                          & Precision             & -                                 &              \textbf{67.91}    &             61.24           &     67.91   \\
                     & Recall                & - & 75.26 & \textbf{81.44} & 75.26 \\
                     & F1                    & - & \textbf{71.39} & 69.91 & 71.39 \\ \hline
Country              & Accuracy  & - & \textbf{66.00} & 65.00 & 66.00 \\ \hline
Region               & Accuracy  & - & \textbf{3.00} & 2.00 & 3.00 \\ \hline
City                 & Accuracy  & - & \textbf{23.00} & 12.00 & 23.00 \\ \hline
Year                 & Accuracy  & - & 46.00 & \textbf{50.00} & 46.00 \\ \hline
Month                & Accuracy  & - & \textbf{33.00} & 29.00 & 33.00 \\ \hline
Day                 & Accuracy  & - & \textbf{14.00} & 8.00 & 14.00 \\ \hline
\end{tabular}%
}
\caption{Overall performance of the baseline models on \dataset test set (\%). All experiments are based on a single run with a preset random seed.}
\label{tab:results}
\end{table*}

\subsection{Overall Performance}
% \begin{table*}[]
%     \centering
%     \begin{tabular}{c|c|c|c|c|c}
%     \hline
%          Class & Metric & Baseline & T5 w/ Truncation & T5 w/ Knowledge Fusion & Hybrid Model \\
%          \midrule
%          \multirow{ 3}{*}{Perpetrator Mention} & Precision & & & & \\
%          & Recall & & & & \\
%          & F1 & & & & \\
%          \hline
%          Perpetrator Type & Accuracy & & & & \\
%          \hline
%          \multirow{ 6}{*}{Victim Name} & Exact Match Precision & & & & \\
%          & Exact Match Recall & & & & \\
%          & Exact Match F1 & & & & \\
%          & Fuzzy Match Precision & & & & \\
%          & Fuzzy Match Recall & & & & \\
%          & Fuzzy Match F1 & & & & \\
%          \hline
%          Victim Type & Accuracy & & & & \\
%          Victim Sex Type & Accuracy & & & & \\
%          Victim Age Group & Accuracy & & & & \\
%          \multirow{ 3}{*}{Violation Type} & Precision & & & & \\
%          & Recall & & & & \\
%          & F1 & & & & \\
%          \hline
%          Country & Accuracy & & & & \\
%          \hline
%          Region & Accuracy & & & & \\
%          \hline
%          City & Accuracy & & & & \\
%          \hline
%          Year & Accuracy & & & & \\
%          \hline
%          Month & Accuracy & & & & \\
%          \hline
%          Day & Accuracy & & & & \\
%     \hline
%     \end{tabular}
%     \caption{Caption}
%     \label{tab:results}
% \end{table*}

Table~\ref{tab:results} shows the performance of the four models on the test set: the DyGIE++ baseline, T5 w/ Truncation, T5 w/ Knowledge Fusion, and the Hybrid model.
%\joel{This links to my earlier comment about baselines. What are baselines and proposed models?  This needs to be defined.   } \shihao{Is this better? All of them are baseline models, also cycles back to the idea that we are not proposing new models, simply baseline models.} 
Both T5-based models significantly outperform the DyGIE++ baseline, except for the precision of \textsc{Perpetrator Mention}. In addition,  we get further improvement from the Knowledge Fusion method for the \textsc{Perpetrator Mention}, \textsc{Victim Name}, and \textsc{Year} attributes. For other attributes, we get results that are slightly worse than those without Knowledge Fusion. This aligns with our assumption that violation events may be elaborated in the later parts of the news articles with specific victim names and violation types. So by applying the Knowledge Fusion method, we can significantly improve the recall of some event attributes. But for other information such as violation time and location, they usually appear in the first several sentences of the news article. The time and location information appearing in the later parts may not be related to the primary attacking event. So based on the evaluation results on the dev set (Table~\ref{tab:results_dev} in Appendix~\ref{appendix:dev}), we propose a hybrid model as our final baseline model. The hybrid model only applies Knowledge Fusion to \textsc{Perpetrator Mention}, \textsc{Victim Name}, and \textsc{Victim Age Group} attributes. We notice that the hybrid model designed based on the dev set does not achieve the best performance for \textsc{Victim Age Group} and \textsc{Year} attributes on the test set. It might be the fact that the hybrid strategy is overfitted on the dev set. And we leave the optimization of the hybrid model as future work.

While the hybrid model outperforms the DyGIE++ baseline in almost all of the event attributes and unlocks the extraction of new attributes, we do see a relatively lower model performance in attributes such as \textsc{Region} and \textsc{Day}. We hypothesize that the ambiguity in \textsc{Region} labels and the large number of classes in \textsc{Day} labels introduce additional challenges to the model, especially with a limited amount of training data. For instance, some annotators mistakenly put \textit{London} under \textsc{Region} instead of \textsc{City}. We acknowledge that the annotation instructions could be further improved to address this issue.

\begin{table}[!ht]
\centering
% \resizebox{\columnwidth}{!}{%
\begin{tabular}{llc}
\hline
\rowcolor[HTML]{ECF4FF}
\cellcolor[HTML]{ECF4FF}Event Attribute                                          & \cellcolor[HTML]{ECF4FF}Metric                & Hybrid \\ \hline
\midrule
Victim Type          & F1              &  22.89\\ \hline
Victim Sex Type      & F1              &  33.33 \\ \hline
Victim Age Group     & F1              &  46.01\\ \hline
\end{tabular}%
% }
\caption{End-to-end performance of the Hybrid model on \dataset (\%) for victim-dependent attributes with model predicted victim names. All experiments are based on a single run with a preset random seed.}
\label{tab:e2e}
\end{table}

We also evaluate the end-to-end performance on the victim-dependent attributes with the model-predicted victim names (Table~\ref{tab:e2e}). And we use F1 scores as the evaluation metric. One victim-dependent attribute is counted as correct only when both the predicted victim name and the predicted attribute value match with the ground truth. 
% \joel{there should be discussion on the difficult areas for the model to get right, and why they are challenging.  }

\section{Conclusion}
%\aoife{I wonder if the emphasis should be on the dataset and not the task, since it's a rather straightforward prediction task?}
%\joel{I agree with Aoife, though technically a new task is introduced by virtue of the dataset}
%\shihao{Rewrote}
In this paper, we present a new dataset that supports extracting detailed information about attacks on human rights defenders under a new task setting. Compared with existing event extraction resources, we focus on the human rights domain and expand to more event attributes for capturing event details more comprehensively. Our new dataset (\dataset) contains 500 human-annotated news articles with 13 different event attributes regarding the victim(s), the type of perpetrator and violation(s), as well as the time and location of the attacks. We demonstrate the usefulness of the dataset by developing a Sequence-to-Sequence-based Question Answering model tailored for this task.  While it achieves decent performance on some event attributes, there are many where there is much room for improvement.  We view this model as a strong baseline for future work.  
%to a pre-trained off-the-shelf Event Extraction model. The models trained on \dataset show marginal improvement on all the sub-tasks to predict the annotated characteristics of the attack events. 
We believe models trained with \dataset could be generalized to detect attacking events in other domains or targeting a different population. And we hope that this work encourages additional research on the development of new AI4SG
%wish to pave a road for unlocking more AI4SG %\aoife{This is the only time this acronym appears in the paper}\shihao{Added another entry in intro} 
NLP resources in the future.

%\clearpage
\section*{Acknowledgements}
We would like to thank Jessie End at Dataminr for her support during this project. We also want to thank all the reviewers for their valuable and constructive feedback during the review phase. 

\section*{Limitations}
While \dataset is, to the best of our knowledge, the first dataset on extracting attacks on human rights defenders, there are some limitations. For one, while being the first corpus of its kind, our dataset is English-only. Second, the number of documents is limited. While the sample size of \dataset (500) is on par with some of the other EE datasets, such as ACE05 (599), we do see more samples being beneficial to subsequent model training and supporting other future studies. In addition, despite the effort to balance the class labels in the event attributes, some of the labels still remain imbalanced, such as \textsc{Perpetrator Type}.

% \joel{other limitation: it's only in English but note that this is the first corpus of its kind.}

\section*{Ethics Statement}
The construction of \dataset involves human annotations on AMT. The Turkers are provided with clear annotation instructions and are informed of the conditions where they would be qualified or disqualified. We compensate the Turkers with a final paid rate of \$15.00 per hour which is over the US national minimal wage of \$7.50. 
%We are also aware that our work could be potentially misused as training samples, though a small number, to develop models for event tracking purposes.
\bibliography{custom}
\bibliographystyle{acl_natbib}

\appendix

\section{Annotation Interface}
\label{appx:annotation_ui}
\begin{figure*}
  \includegraphics[width=\textwidth]{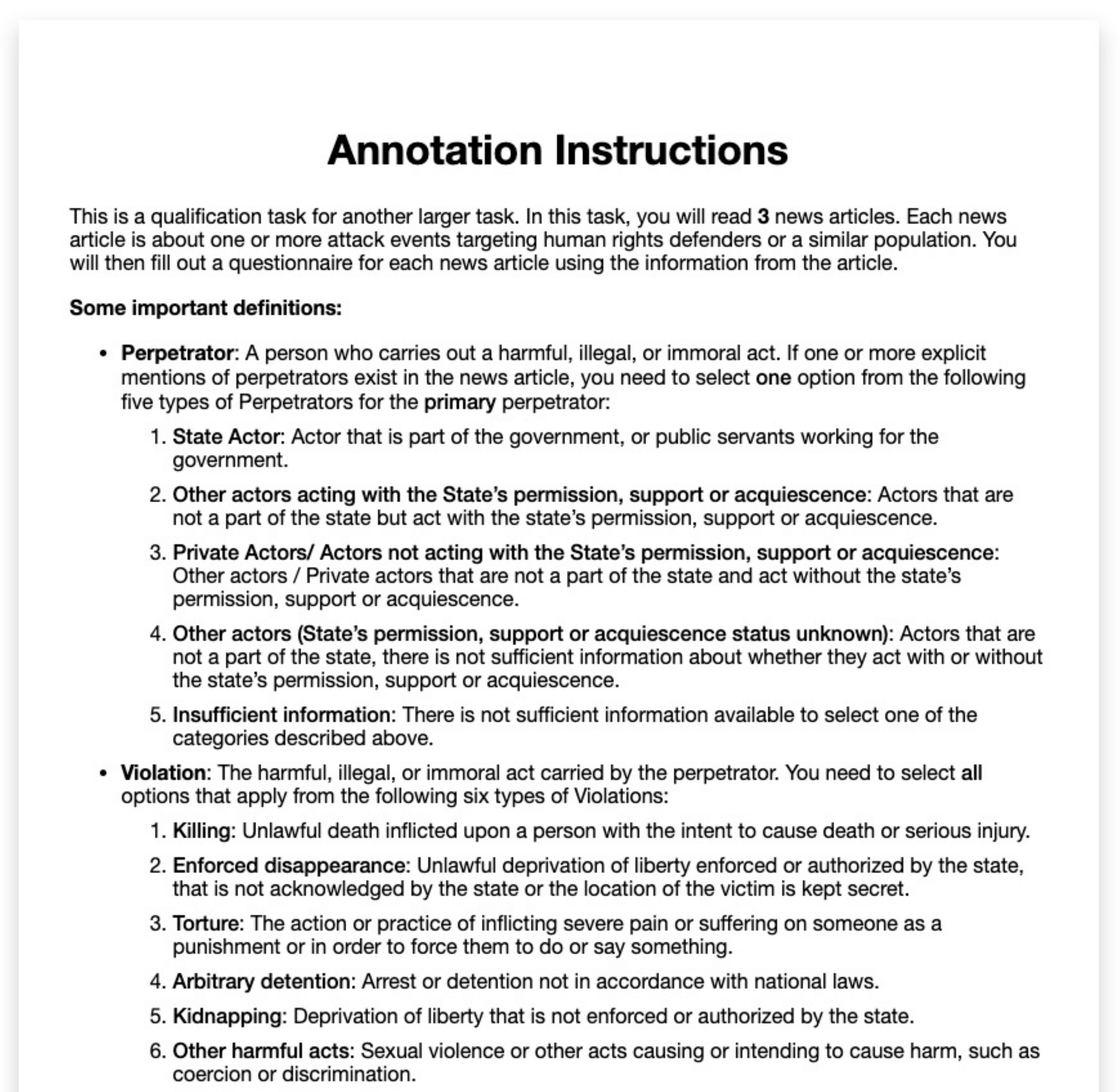}
  \caption{Screenshot of the Qualification Task Instructions (1/3).}
  \label{fig:qua_1}
\end{figure*}

\begin{figure*}
  \includegraphics[width=\textwidth]{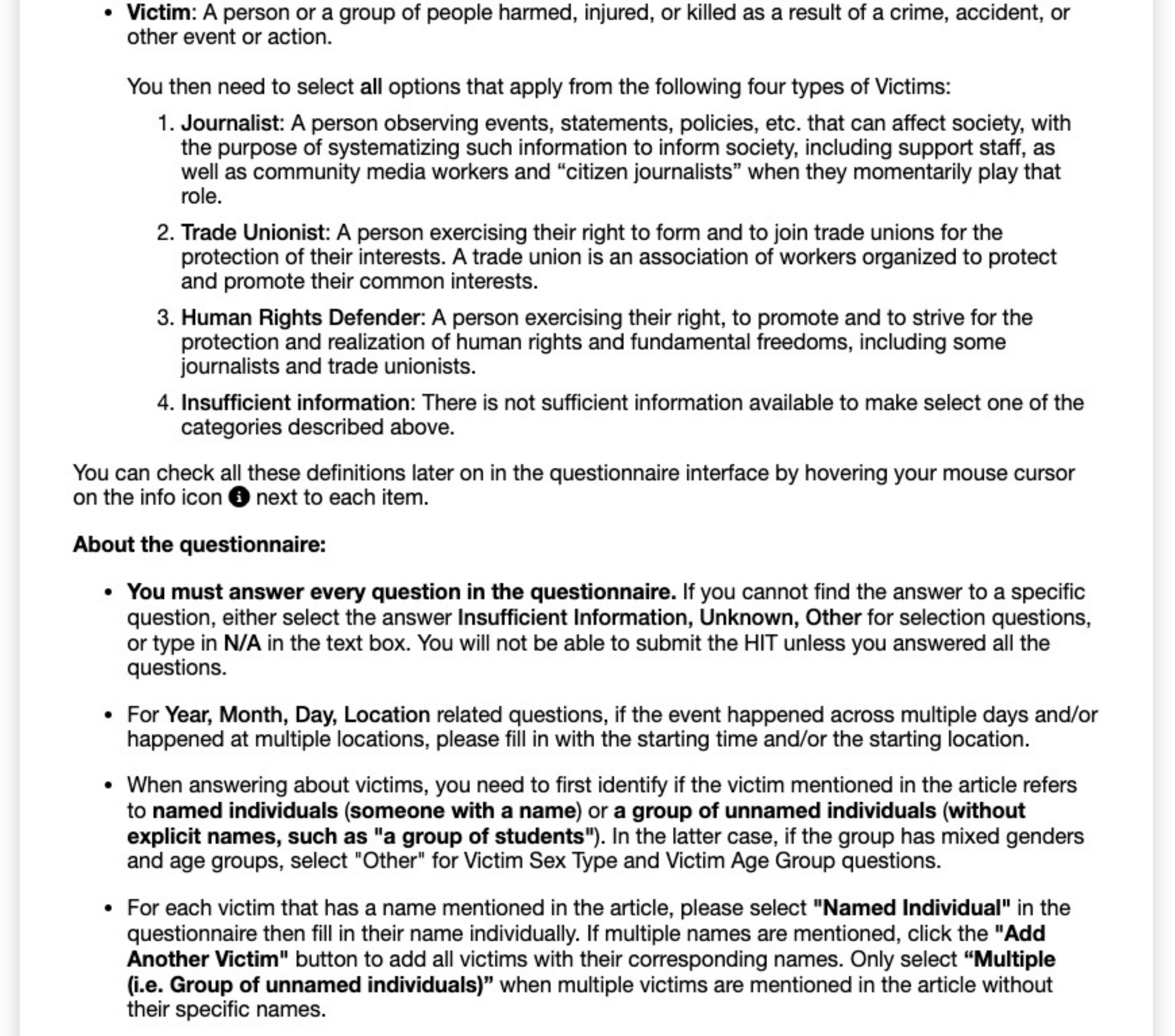}
  \caption{Screenshot of the Qualification Task Instructions (2/3).}
  \label{fig:qua_2}
\end{figure*}

\begin{figure*}
  \includegraphics[width=\textwidth]{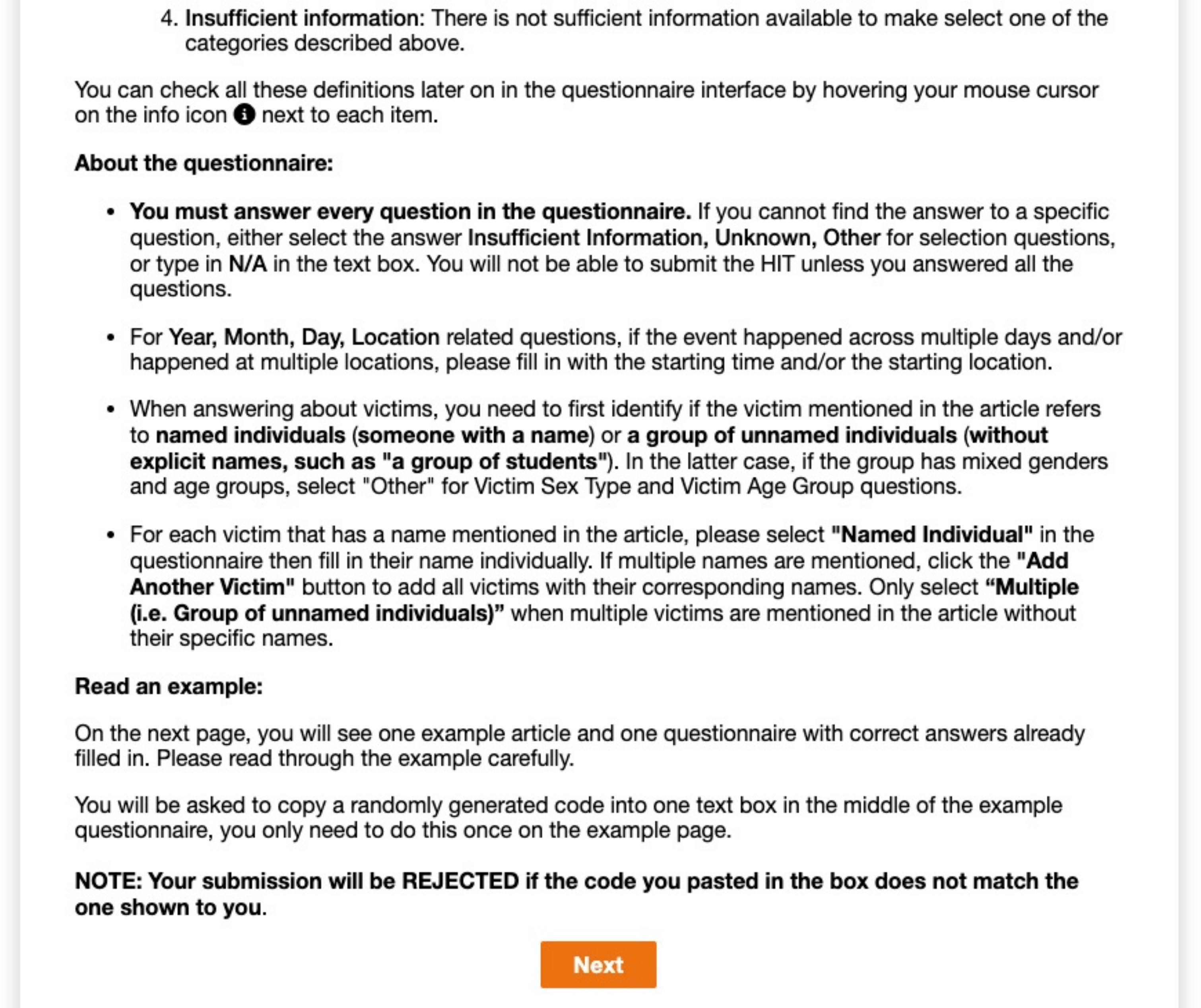}
  \caption{Screenshot of the Qualification Task Instructions (3/3).}
  \label{fig:qua_3}
\end{figure*}

\begin{figure*}
  \includegraphics[width=\textwidth]{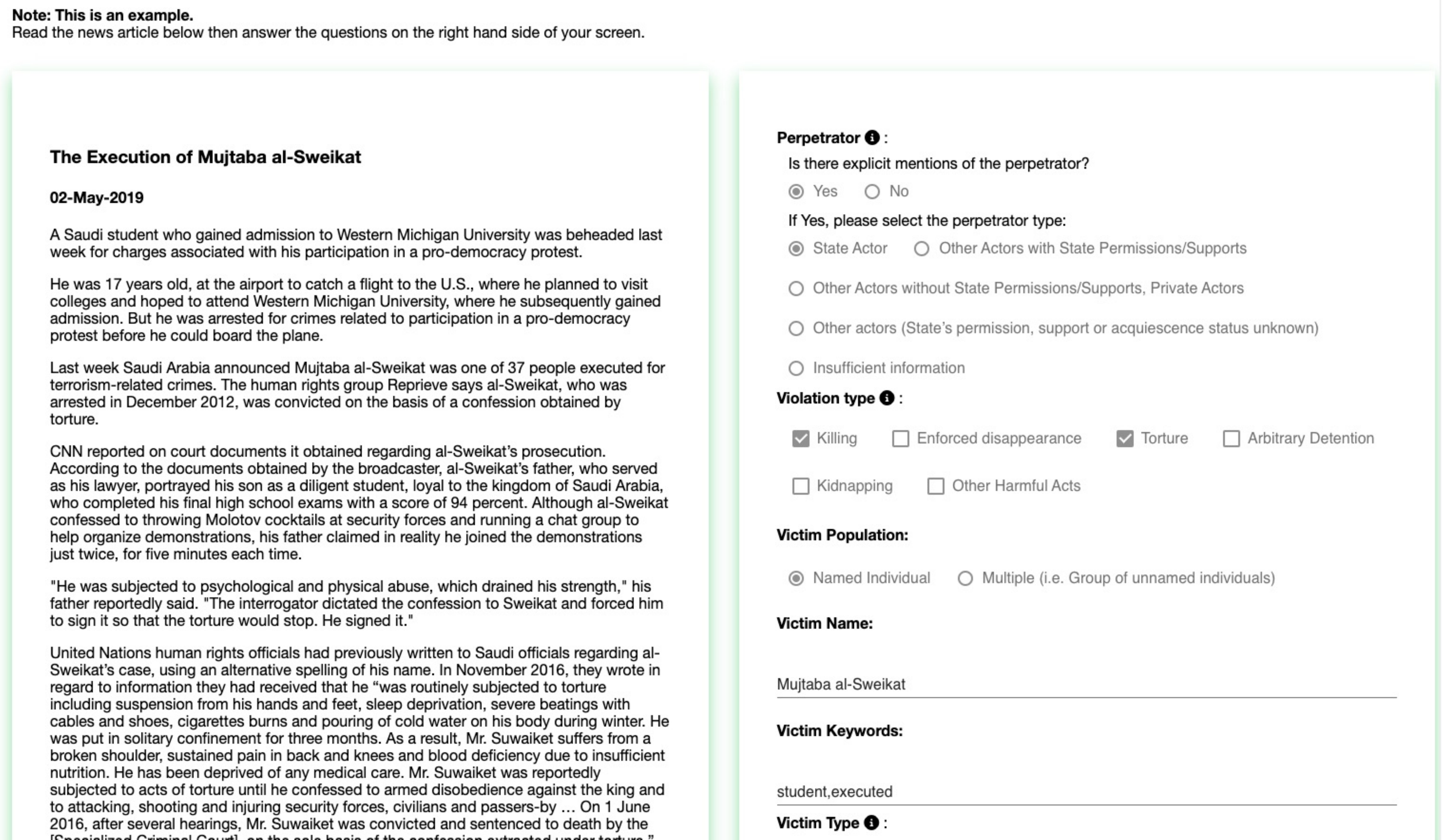}
  \caption{Screenshot of the Qualification Task Example Page (1/2).}
  \label{fig:qua_4}
\end{figure*}

\begin{figure*}
  \includegraphics[width=\textwidth]{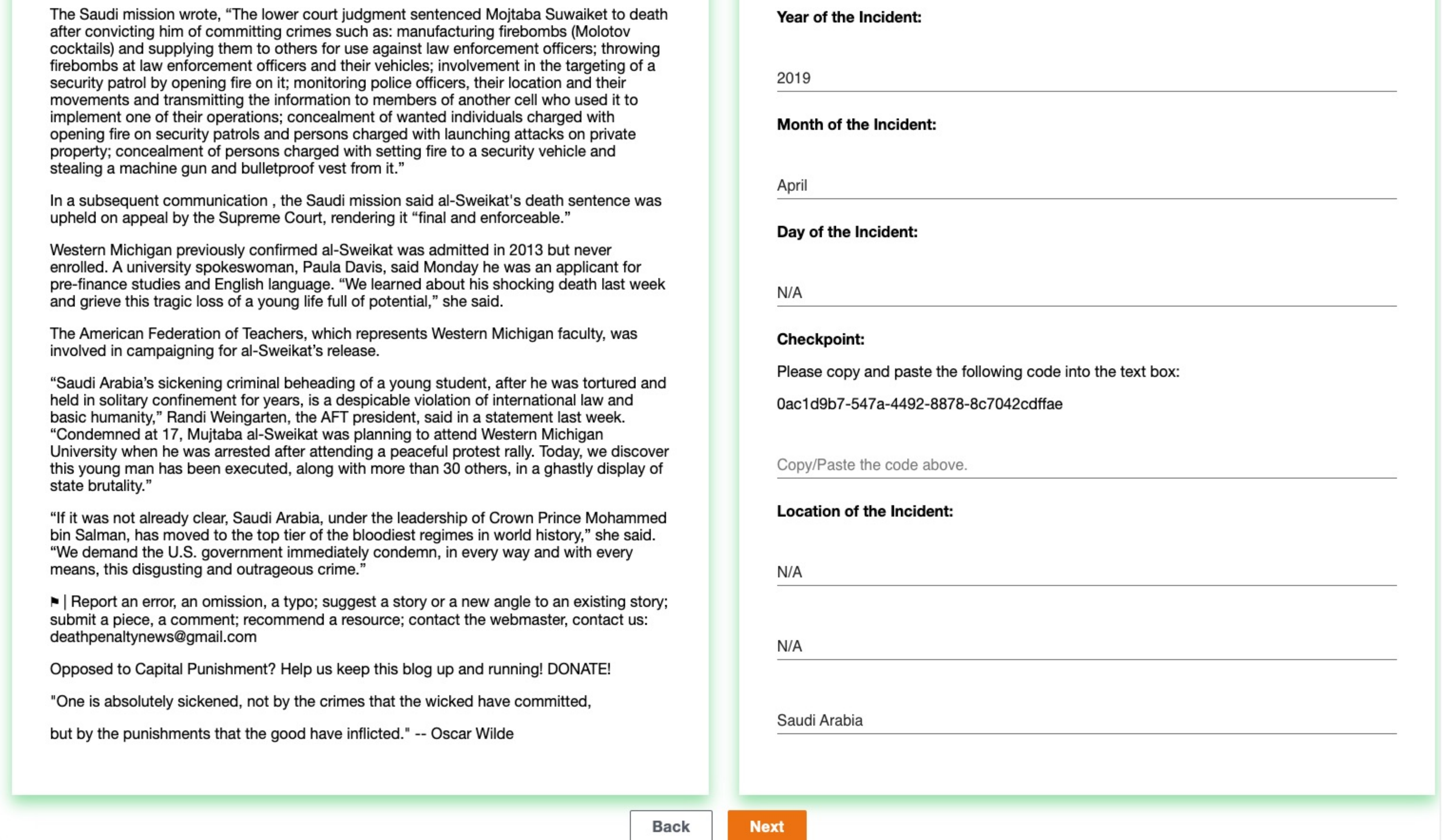}
  \caption{Screenshot of the Qualification Task Example Page (2/2).}
  \label{fig:qua_5}
\end{figure*}

\begin{figure*}
  \includegraphics[width=\textwidth]{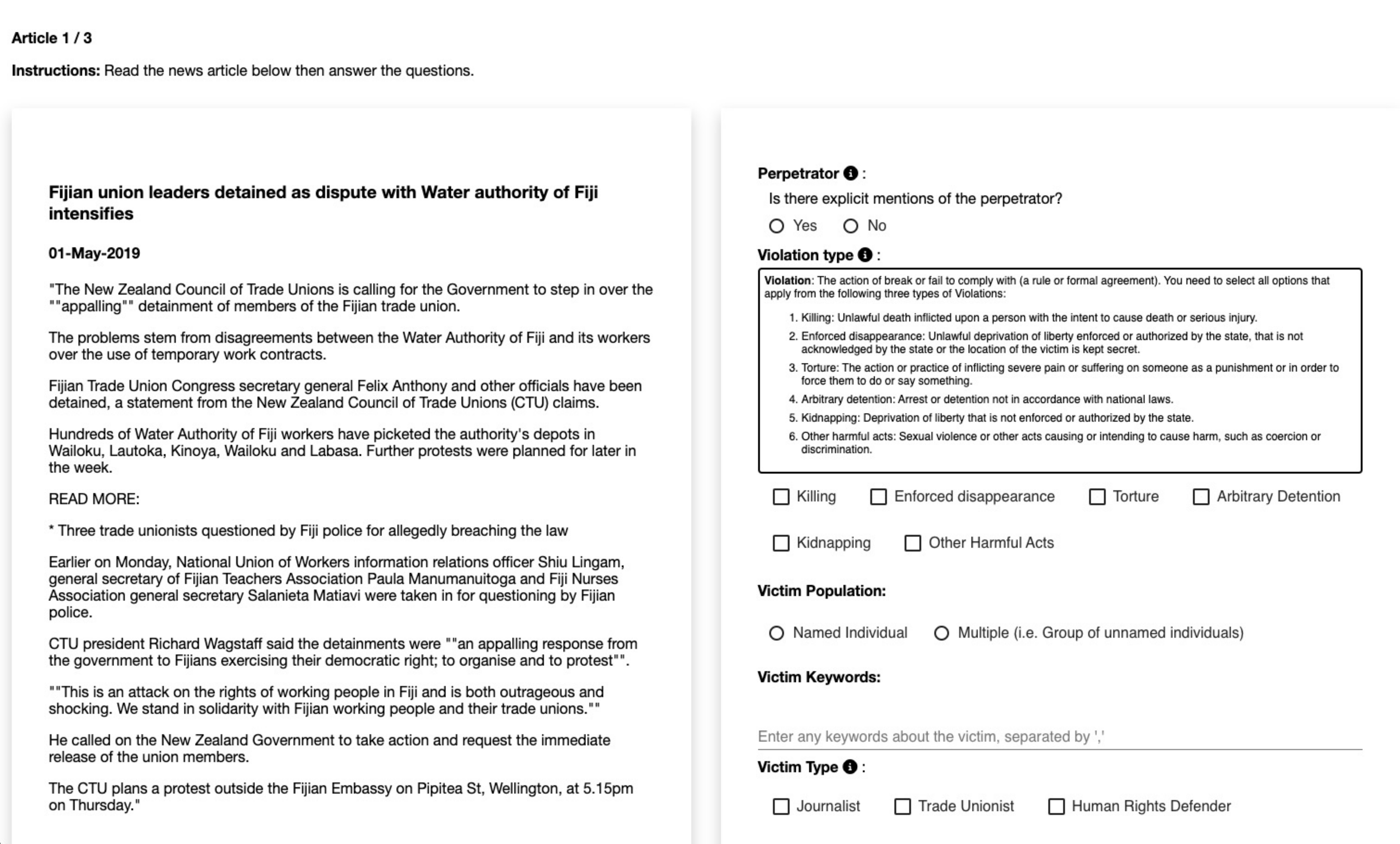}
  \caption{Screenshot of the Qualification Task Articles (1/3). By hovering over the information icon next to \textbf{Violation Type}, Turkers can check the definitions of all violation types on this page.}
  \label{fig:qua_6}
\end{figure*}

\begin{figure*}
  \includegraphics[width=\textwidth]{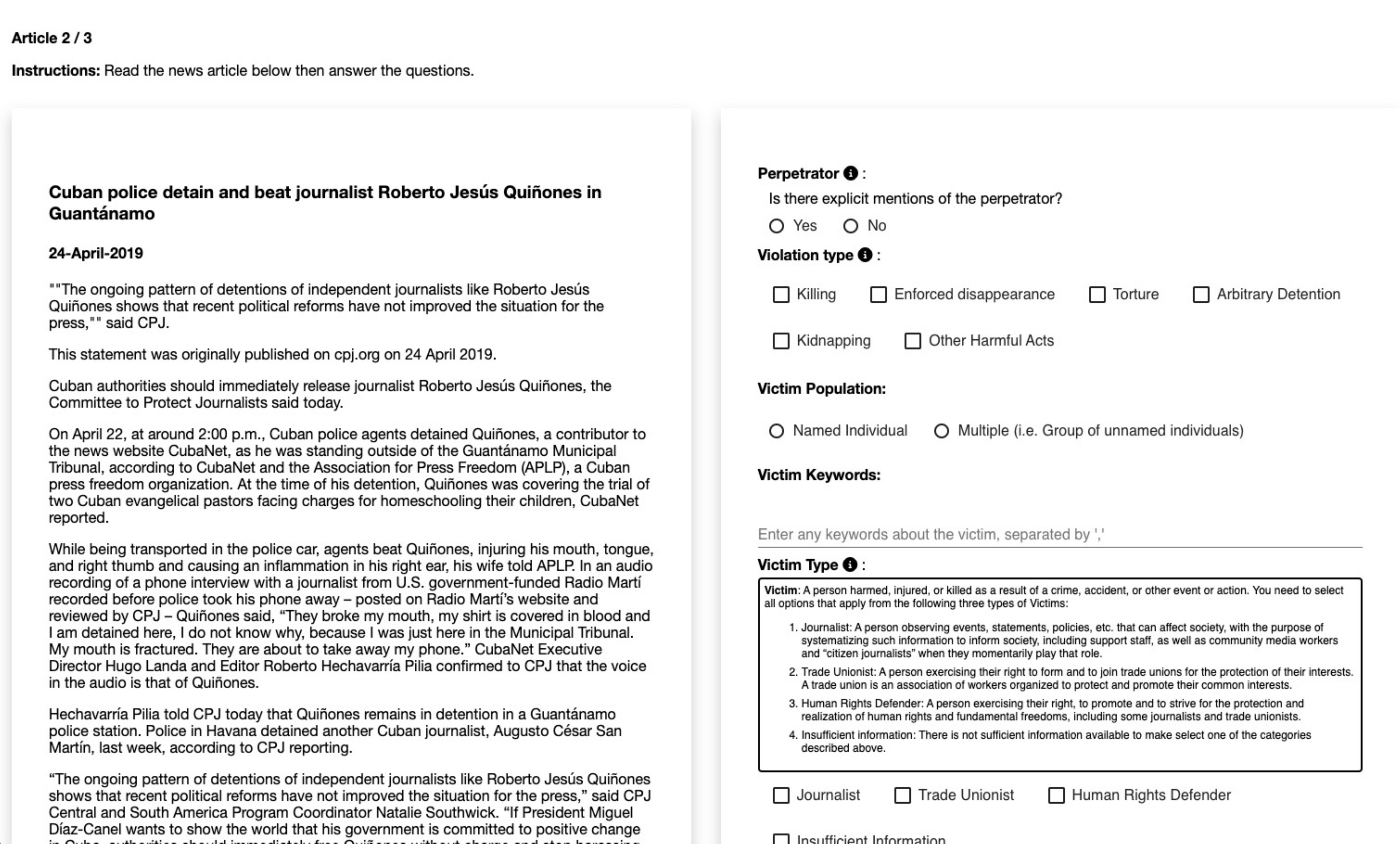}
  \caption{Screenshot of the Qualification Task Articles (2/3). By hovering over the information icon next to \textbf{Victim Type}, Turkers can check the definitions of all victim types on this page.}
  \label{fig:qua_7}
\end{figure*}

\begin{figure*}
  \includegraphics[width=\textwidth]{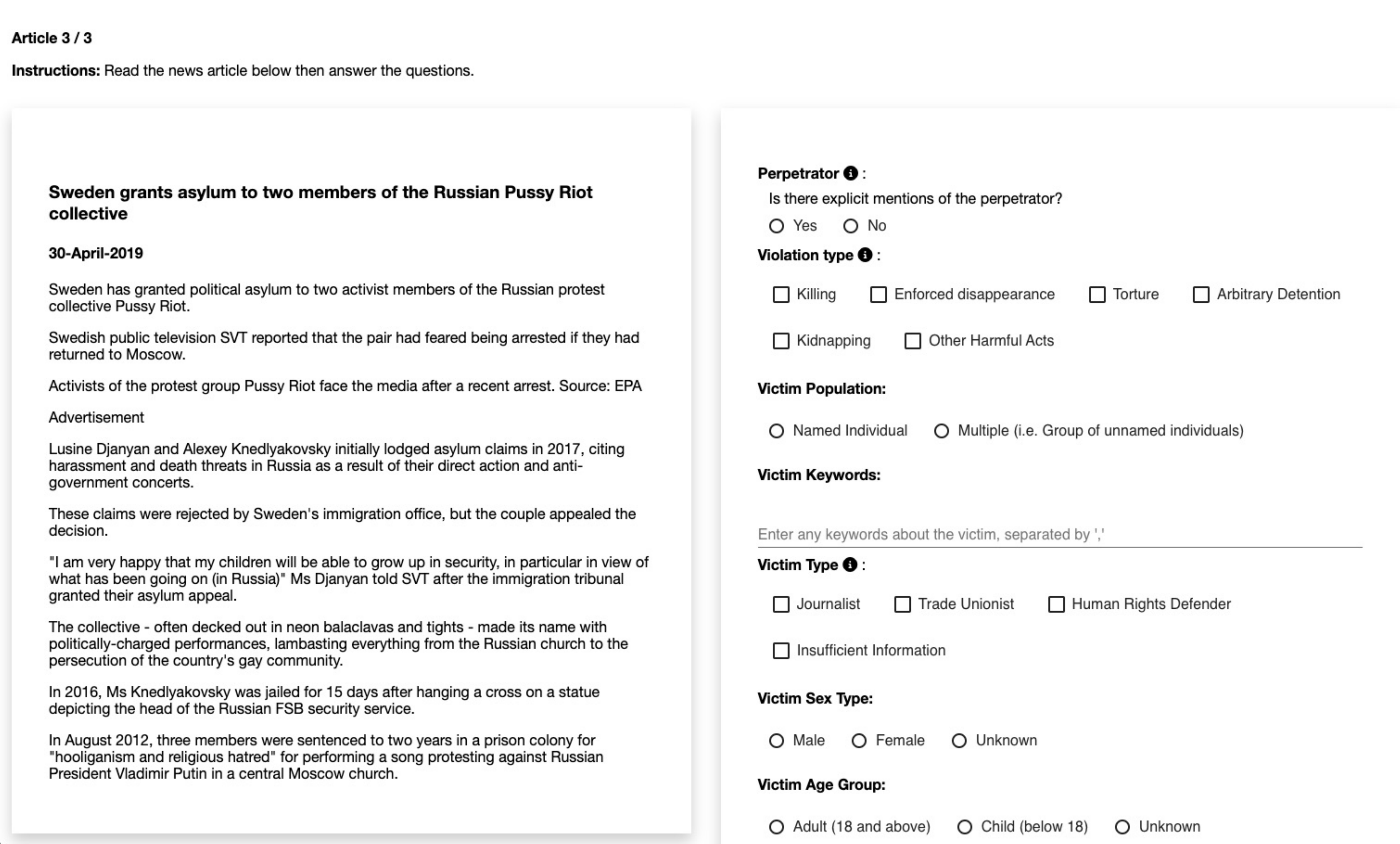}
  \caption{Screenshot of the Qualification Task Articles (3/3).}
  \label{fig:qua_8}
\end{figure*}

\begin{figure*}
  \includegraphics[width=\textwidth]{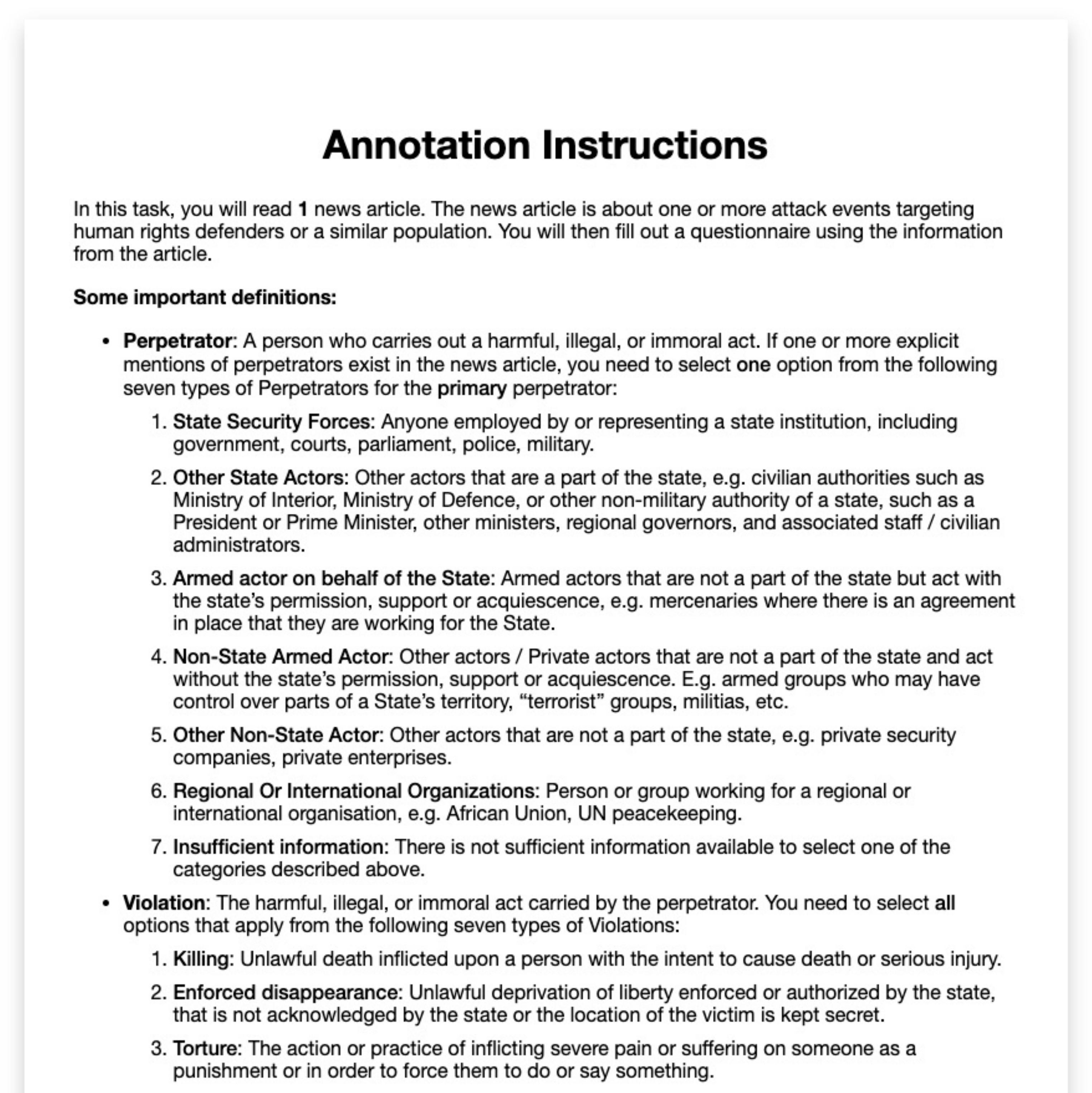}
  \caption{Screenshot of the Full Task Instructions (1/2).}
  \label{fig:full_1}
\end{figure*}

\begin{figure*}
  \includegraphics[width=\textwidth]{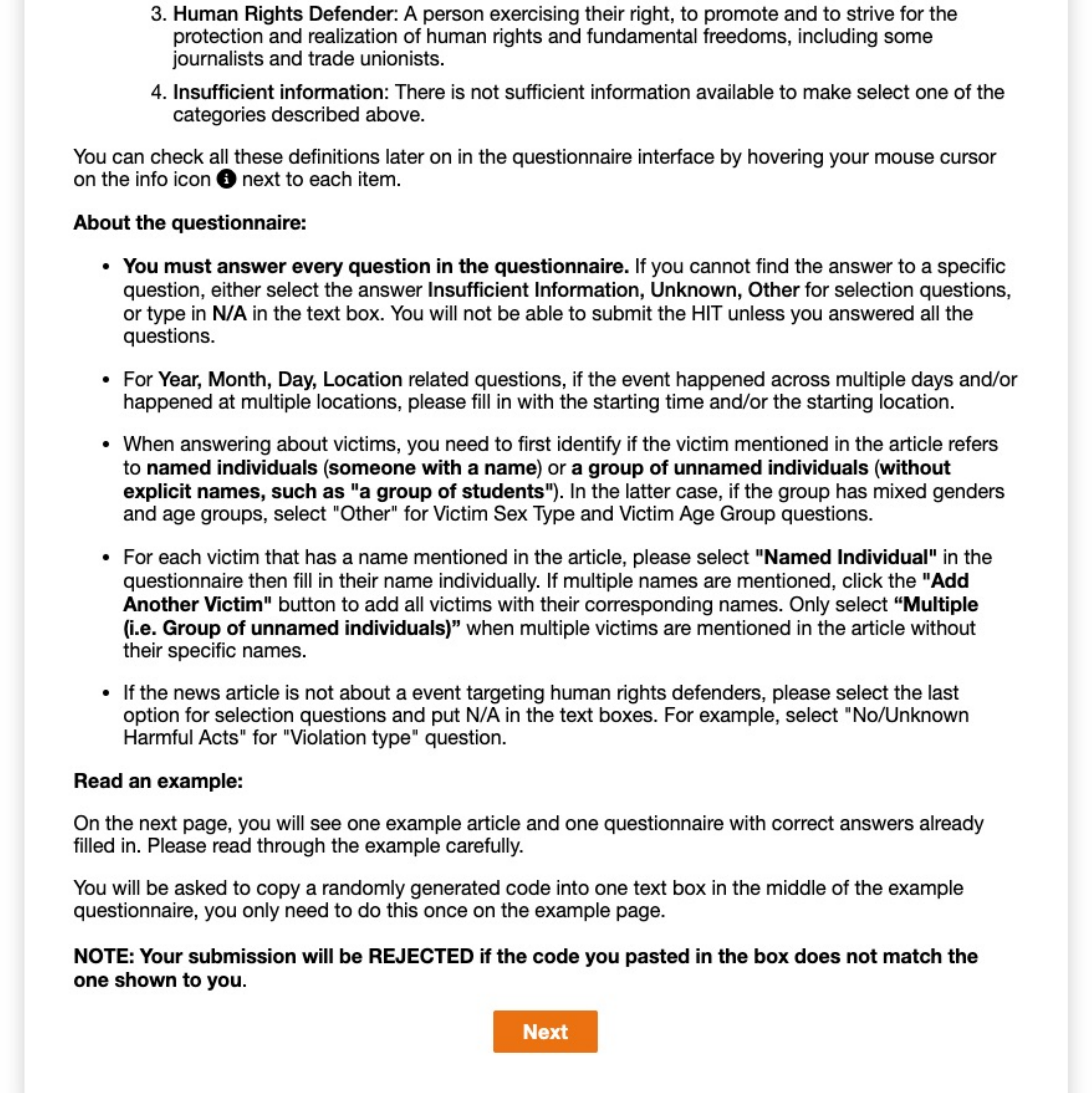}
  \caption{Screenshot of the Full Task Instructions (2/2).}
  \label{fig:full_2}
\end{figure*}

\begin{figure*}
  \includegraphics[width=\textwidth]{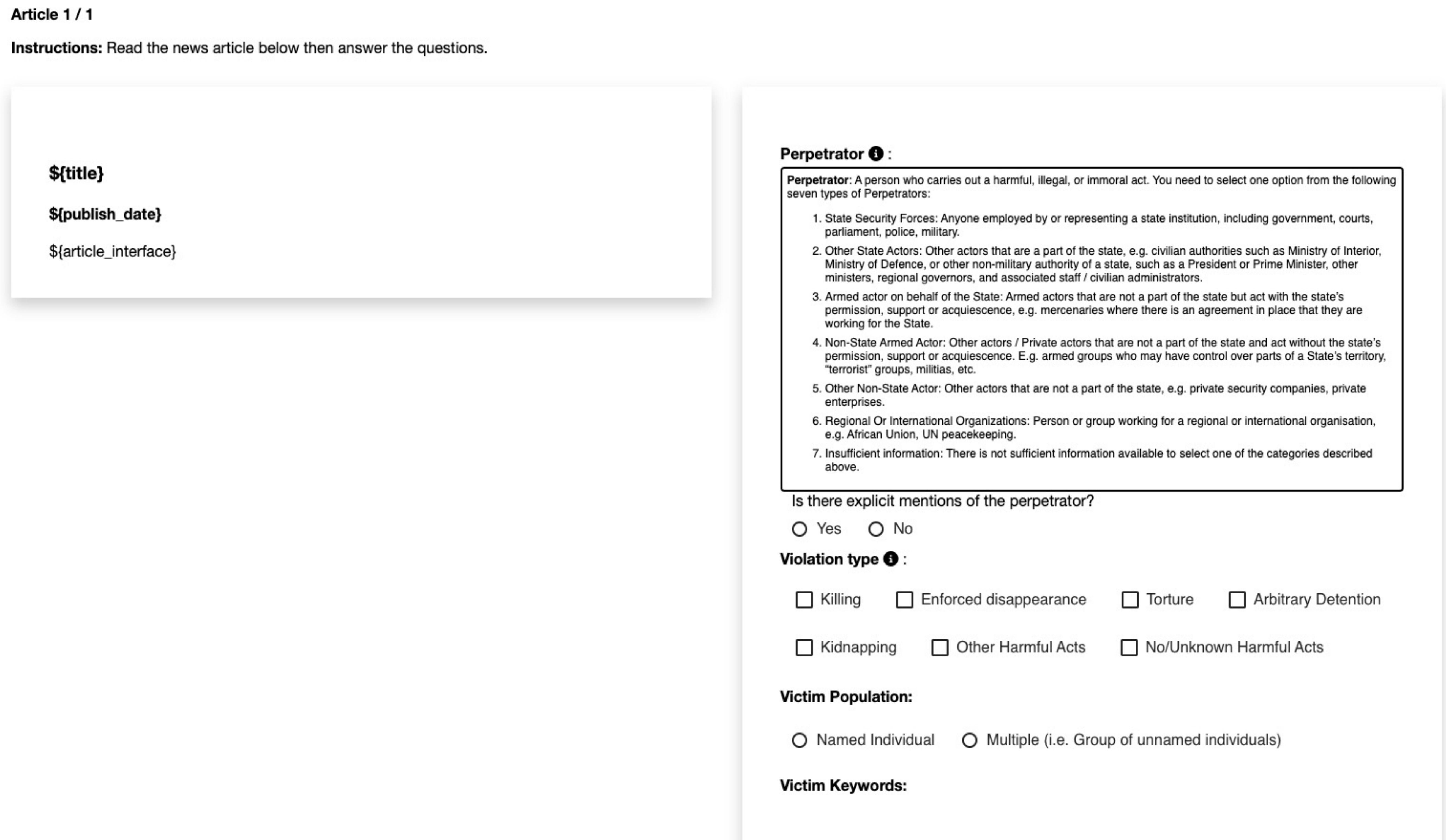}
  \caption{Screenshot of the Full Task Annotation Page (1/2). By hovering over the information icon next to \textbf{Perpetrator}, Turkers can check the definitions of all perpetrator types on this page.}
  \label{fig:full_3}
\end{figure*}

\begin{figure*}
  \includegraphics[width=\textwidth]{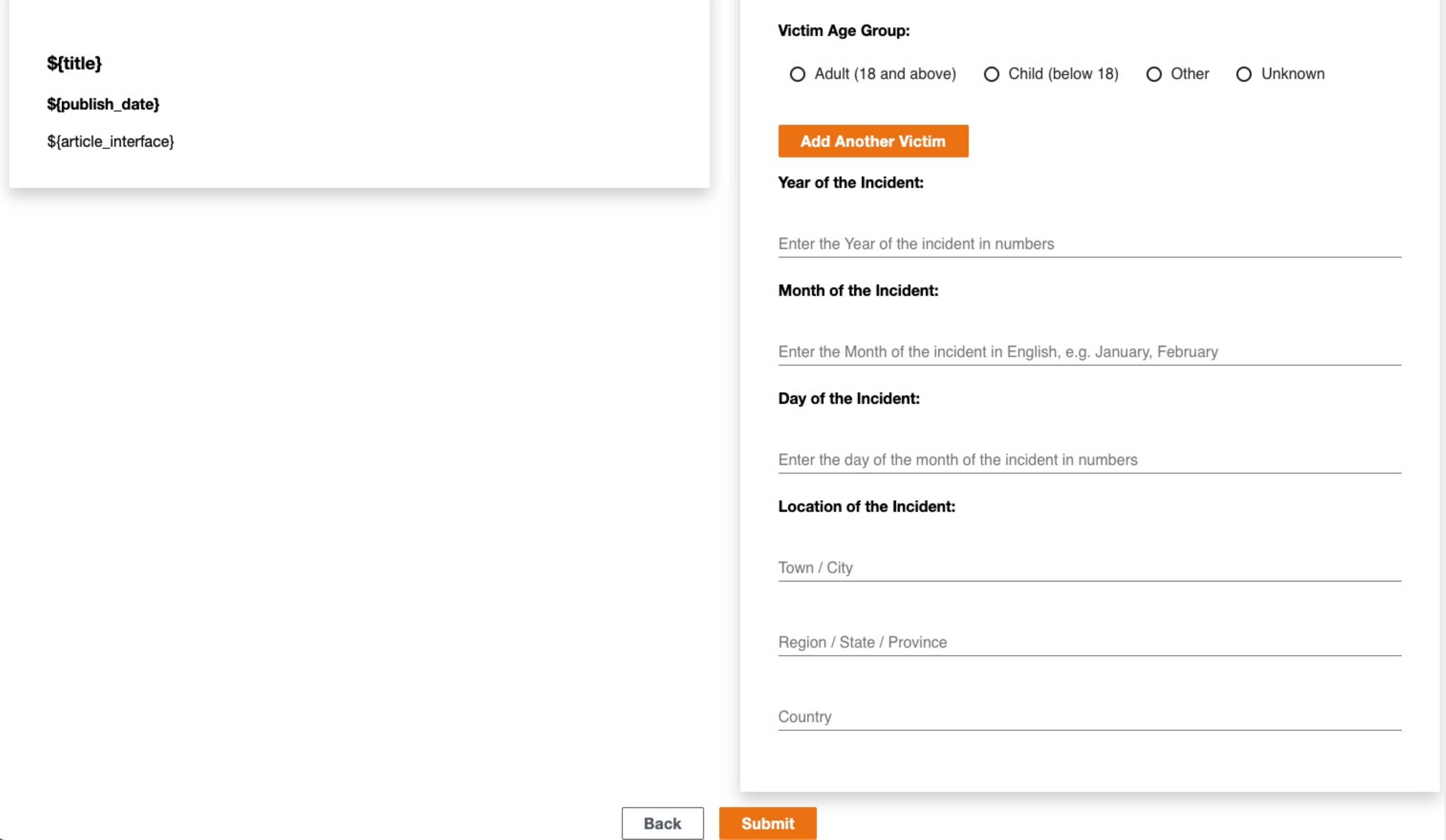}
  \caption{Screenshot of the Full Task Annotation Page (2/2).}
  \label{fig:full_4}
\end{figure*}

We list the screenshots of our annotation interface from Figure~\ref{fig:qua_1} through Figure~\ref{fig:full_4}, including the annotation guideline (instructions) page for both the qualification tasks and the full tasks, the example page, as well as the task pages.

\section{Label Statistics}
\label{appx:label_stats}
In Table~\ref{tab:label_stats}, we list the statistics of all the labels in \dataset, as well as their distributions in the train, dev, and test set.

\begin{table*}[ht]
\centering
\resizebox{\textwidth}{!}{%
\begin{tabular}{|l|l|l|c|c|c|c|}
\hline
\rowcolor[HTML]{ECF4FF} 
\multicolumn{1}{|c|}{\cellcolor[HTML]{ECF4FF}Category} & \multicolumn{1}{c|}{\cellcolor[HTML]{ECF4FF}Event Attribute} & \multicolumn{1}{c|}{\cellcolor[HTML]{ECF4FF}Labels} & \multicolumn{1}{l|}{\cellcolor[HTML]{ECF4FF}Train} & \multicolumn{1}{l|}{\cellcolor[HTML]{ECF4FF}Dev} & \multicolumn{1}{l|}{\cellcolor[HTML]{ECF4FF}Test} & \multicolumn{1}{l|}{\cellcolor[HTML]{ECF4FF}Total} \\ \hline
\midrule
 &  & Yes & 272 & 95 & 91 & 458 \\ \cline{3-7} 
 & \multirow{-2}{*}{Perpetrator Mention} & No & 28 & 5 & 9 & 42 \\ \cline{2-7} 
 &  & State Security Forces & 149 & 60 & 56 & 265 \\ \cline{3-7} 
 &  & Other State Actors & 25 & 6 & 10 & 41 \\ \cline{3-7} 
 &  & Other non-state actors & 34 & 11 & 9 & 54 \\ \cline{3-7} 
 &  & Other actors with permissions & 10 & 5 & 3 & 18 \\ \cline{3-7} 
 &  & Other actors without permissions & 41 & 10 & 10 & 61 \\ \cline{3-7} 
 &  & Regional Organizations & 4 & 1 & 1 & 6 \\ \cline{3-7} 
 &  & Insufficient Information & 9 & 1 & 1 & 11 \\ \cline{3-7} 
\multirow{-10}{*}{Perpetrator} & \multirow{-8}{*}{Perpetrator Type} & None & 28 & 6 & 10 & 44 \\ \hline
 &  & Arbitrary Detention & 138 & 53 & 55 & 246 \\ \cline{3-7} 
 &  & Enforced Disappearance & 27 & 8 & 8 & 43 \\ \cline{3-7} 
 &  & Killing & 109 & 33 & 34 & 176 \\ \cline{3-7} 
 &  & Kidnapping & 76 & 21 & 16 & 113 \\ \cline{3-7} 
 &  & Torture & 56 & 19 & 24 & 99 \\ \cline{3-7} 
 &  & Other & 131 & 46 & 50 & 227 \\ \cline{3-7} 
\multirow{-7}{*}{Violation} & \multirow{-7}{*}{Violation Type} & Unknown & 20 & 5 & 10 & 35 \\ \hline
 & Victim  Name & - & 463 & 198 & 130 & 791 \\ \cline{2-7} 
 &  & Human Rights Defender & 145 & 32 & 42 & 219 \\ \cline{3-7} 
 &  & Trade Unionist & 59 & 25 & 9 & 93 \\ \cline{3-7} 
 &  & Journalist & 195 & 104 & 66 & 365 \\ \cline{3-7} 
 & \multirow{-4}{*}{Victim Type} & Insufficient Information & 356 & 120 & 113 & 589 \\ \cline{2-7} 
 &  & Individual & 463 & 198 & 130 & 791 \\ \cline{3-7} 
 & \multirow{-2}{*}{Victim Population Type} & Multiple & 224 & 74 & 74 & 372 \\ \cline{2-7} 
 &  & Adult & 491 & 217 & 131 & 839 \\ \cline{3-7} 
 &  & Child & 19 & 5 & 9 & 33 \\ \cline{3-7} 
 &  & Other & 34 & 4 & 17 & 55 \\ \cline{3-7} 
 & \multirow{-4}{*}{Victim Age Group} & Unknown & 143 & 46 & 47 & 236 \\ \cline{2-7} 
 &  & Man & 274 & 115 & 90 & 479 \\ \cline{3-7} 
 &  & Woman & 115 & 35 & 33 & 183 \\ \cline{3-7} 
 &  & Other & 76 & 15 & 29 & 120 \\ \cline{3-7} 
\multirow{-15}{*}{Victim} & \multirow{-4}{*}{Victim Sex Group} & Unknown & 222 & 107 & 52 & 381 \\ \hline
 & Country & - & 279 & 89 & 89 & 457 \\ \cline{2-7} 
 & Region & - & 69 & 26 & 12 & 107 \\ \cline{2-7} 
\multirow{-3}{*}{Location} & City & - & 163 & 53 & 45 & 261 \\ \hline
 & Year & - & 268 & 83 & 80 & 431 \\ \cline{2-7} 
 & Month & January, ..., December & 183 & 60 & 48 & 291 \\ \cline{2-7} 
\multirow{-3}{*}{Time} & Day & 1, 2, 3, ..., 31 & 109 & 35 & 35 & 179 \\ \hline
\end{tabular}%
}
\caption{Label statistics of \dataset.}
\label{tab:label_stats}
\end{table*}

\section{Cohen-Kappa Scores}
\label{appx:kappa_score}
We calculate the average pair-wise Cohen-Kappa scores for each qualified Turker during our pilot study using 100 Hits with replication factor 3. While we did our due diligence to make our annotation instructions as comprehensive as possible, some of the concepts regarding Human Rights were sometimes challenging to distinguish for the Turkers. The relatively lower weighted average of Cohen-Kappa scores for some event attributes (\textsc{Perpetrator Mention}: $0.40$, \textsc{Perpetrator Type}: $0.41$) are also potentially due to the imbalanced distributions of these attributes. The weighted averages of Cohen-Kappa scores for other attributes are all higher than $0.61$ for violation and victim-related classes (\textsc{Violation Type}: $0.67$, \textsc{Victim Population Type}: $0.64$, \textsc{Victim Type}: $0.62$), which are generally considered as substantially agree. 

\begin{table*}[ht]
\centering
\resizebox{\textwidth}{!}{%
\begin{tabular}{|c|c|ccccc|c|}
\hline
\rowcolor[HTML]{ECF4FF} 
\cellcolor[HTML]{ECF4FF} &
  \cellcolor[HTML]{ECF4FF} &
  \multicolumn{5}{c|}{\cellcolor[HTML]{ECF4FF}Average Pair-wise Cohen-Kappa Score} &
  \cellcolor[HTML]{ECF4FF} \\ \cline{3-7}
\rowcolor[HTML]{ECF4FF} 
\multirow{-2}{*}{\cellcolor[HTML]{ECF4FF}\begin{tabular}[c]{@{}c@{}}Worker\\ Index\end{tabular}} &
  \multirow{-2}{*}{\cellcolor[HTML]{ECF4FF}\begin{tabular}[c]{@{}c@{}}No. of HITs \\ Finished\end{tabular}} &
  \multicolumn{1}{c|}{\cellcolor[HTML]{ECF4FF}Perpetrator Mention} &
  \multicolumn{1}{c|}{\cellcolor[HTML]{ECF4FF}Perpetrator Type} &
  \multicolumn{1}{c|}{\cellcolor[HTML]{ECF4FF}Violation Type} &
  \multicolumn{1}{c|}{\cellcolor[HTML]{ECF4FF}Victim Population Type} &
  Victim Type &
  \multirow{-2}{*}{\cellcolor[HTML]{ECF4FF}\begin{tabular}[c]{@{}c@{}}Worker\\ Average\end{tabular}} \\ \hline
1 &
  85 &
  \multicolumn{1}{c|}{0.48} &
  \multicolumn{1}{c|}{0.41} &
  \multicolumn{1}{c|}{0.74} &
  \multicolumn{1}{c|}{0.65} &
  0.66 &
  0.59 \\ \hline
2 &
  51 &
  \multicolumn{1}{c|}{0.22} &
  \multicolumn{1}{c|}{0.43} &
  \multicolumn{1}{c|}{0.63} &
  \multicolumn{1}{c|}{0.53} &
  0.61 &
  0.48 \\ \hline
3 &
  51 &
  \multicolumn{1}{c|}{0.60} &
  \multicolumn{1}{c|}{0.53} &
  \multicolumn{1}{c|}{0.74} &
  \multicolumn{1}{c|}{0.82} &
  0.66 &
  0.67 \\ \hline
4 &
  47 &
  \multicolumn{1}{c|}{-0.04} &
  \multicolumn{1}{c|}{0.31} &
  \multicolumn{1}{c|}{0.72} &
  \multicolumn{1}{c|}{0.74} &
  0.57 &
  0.46 \\ \hline
5 &
  38 &
  \multicolumn{1}{c|}{0.45} &
  \multicolumn{1}{c|}{0.37} &
  \multicolumn{1}{c|}{0.70} &
  \multicolumn{1}{c|}{0.46} &
  0.63 &
  0.52 \\ \hline
6 &
  15 &
  \multicolumn{1}{c|}{0.71} &
  \multicolumn{1}{c|}{0.49} &
  \multicolumn{1}{c|}{0.45} &
  \multicolumn{1}{c|}{0.71} &
  0.38 &
  0.55 \\ \hline
7 &
  9 &
  \multicolumn{1}{c|}{1.00} &
  \multicolumn{1}{c|}{0.24} &
  \multicolumn{1}{c|}{0.32} &
  \multicolumn{1}{c|}{0.40} &
  0.80 &
  0.55 \\ \hline
8 &
  2 &
  \multicolumn{1}{c|}{0.50} &
  \multicolumn{1}{c|}{0.50} &
  \multicolumn{1}{c|}{0.00} &
  \multicolumn{1}{c|}{1.00} &
  0.00 &
  0.40 \\ \hline
9 &
  1 &
  \multicolumn{1}{c|}{0.00} &
  \multicolumn{1}{c|}{0.00} &
  \multicolumn{1}{c|}{0.00} &
  \multicolumn{1}{c|}{0.00} &
  0.00 &
  0.00 \\ \hline
10 &
  1 &
  \multicolumn{1}{c|}{0.00} &
  \multicolumn{1}{c|}{0.00} &
  \multicolumn{1}{c|}{0.00} &
  \multicolumn{1}{c|}{0.00} &
  0.00 &
  0.00 \\ \hline
\multicolumn{2}{|c|}{Weighted Average} &
  \multicolumn{1}{c|}{0.40} &
  \multicolumn{1}{c|}{0.41} &
  \multicolumn{1}{c|}{0.67} &
  \multicolumn{1}{c|}{0.64} &
  0.62 &
  - \\ \hline
\end{tabular}%
}
\caption{Turker agreement scores for some of the event attributes calculated during the pilot study with 100 HITs, replication factor 3. }
\label{tab:worker_agreement}
\end{table*}

\section{Input-Output Design}
\label{appx:in_output}
Table~\ref{tab:general_article_dependent_classes} shows the input questions and answers for all the event attributes covered in \dataset. And Table~\ref{tab:task_prefix} shows all the task prefixes that we add to the beginning of the input text in our multi-task training regime. 

\begin{table*}[t]
\centering
\resizebox{\textwidth}{!}{%
\begin{tabular}{l|lll}
\hline
\rowcolor[HTML]{ECF4FF}
\multicolumn{1}{c}{\cellcolor[HTML]{ECF4FF}Category} & \multicolumn{1}{c}{\cellcolor[HTML]{ECF4FF}Event Attribute} & \multicolumn{1}{c}{\cellcolor[HTML]{ECF4FF}Input Question}                          & \multicolumn{1}{c}{\cellcolor[HTML]{ECF4FF}Output Answer} \\ \hline
\midrule
\multirow{12}{*}{\textbf{Gen}} & Perpetrator Mention                & Does it mention any perpetrator?                                     & One of \{Yes, No\}                         \\ \cline{2-4}
& Perpetrator Type &
  What is the type of the perpetrator? &
  \begin{tabular}[c]{@{}l@{}}One of \{state security forces, regional organizations, \\ other actors with permissions, other actors without permissions, \\ other state actors, other non-state actors, insufficient info\}\end{tabular} \\ \cline{2-4}
& \multirow{6}{*}{Violation Type}    & Is there any arbitrary detention violation mentioned in the text?    & \multirow{6}{*}{One of \{Yes, No\}}        \\
&                                   & Is there any enforced disappearance violation mentioned in the text? &                                            \\
&                                   & Is there any kidnapping violation mentioned in the text?             &                                            \\
&                                   & Is there any killing violation mentioned in the text?                &                                            \\
&                                   & Is there any torture violation mentioned in the text?                &                                            \\
&                                   & Is there any other violation mentioned in the text?                  &                                            \\ \cline{2-4}
& Victim Name                        & Who is the victim of the violation?                                 & \{VICTIM\_NAME1, VICTIM\_NAME2, …\}        \\ \cline{2-4}
& Country                            & In which country did the violation happen?                           & \{COUNTRY\_NAME\}                          \\ \cline{2-4}
& Region                             & In which region did the violation happen?                            & \{REGION\_NAME\}                           \\ \cline{2-4}
& City                               & In which city did the violation happen?                              & \{CITY\_NAME\}                             \\ 
\hline
\hline
\multirow{6}{*}{\textbf{Vic}} & Victim Sex Type              & What is the sex of \{VICTIM\_NAME\}?             & One of \{woman, man, other, unknown\}   \\ 
\cline{2-4}
&Victim Age Group             & What is the age group of \{VICTIM\_NAME\}?       & One of \{adult, child, other, unknown\} \\ 
\cline{2-4}
&Victim Population Type       & What is the population type of \{VICTIM\_NAME\}? & One of \{Individual, multiple\}         \\ 
\cline{2-4}
&\multirow{3}{*}{Victim type} & Is \{VICTIM\_NAME\} a trade unionist?            & \multirow{3}{*}{One of \{Yes, No\}}     \\
&                             & Is \{VICTIM\_NAME\} a journalist?                &                                         \\
&                             & Is \{VICTIM\_NAME\} a human rights defender?     &                                         \\
\hline
\hline
\multirow{6}{*}{\textbf{Tim}} & Year  & In which year did the violation happen?      & Year (YYYY)                   \\ 
\cline{2-4}
& Month & In which month did the violation happen?     & Month (month name)            \\ 
\cline{2-4}
& Day  & On which day did the violation happen?      & Day (D with no leading zeros) \\ 
\cline{2-4}
& \multirow{3}{*}{Victim type}       & Is \{VICTIM\_NAME\} a trade unionist?       & \multirow{3}{*}{One of \{Yes, No\}}        \\
&       & Is \{VICTIM\_NAME\} a journalist?            &                               \\
&       & Is \{VICTIM\_NAME\} a human rights defender? &                               \\ 
\hline
\end{tabular}%
}
\caption{Summary of the predefined questions and answers for event attributes. \textbf{Gen} stands for the category of general article-dependent attributes, \textbf{Vic} stands for the category of victim-dependent attributes, and \textbf{Tim} stands for the category of publication time-dependent attributes.}
\label{tab:general_article_dependent_classes}
\end{table*}

\begin{table}[!th]
\centering
\resizebox{\columnwidth}{!}{%
\begin{tabular}{ll}
\hline
\rowcolor[HTML]{ECF4FF}
\multicolumn{1}{c}{\cellcolor[HTML]{ECF4FF}Class} & \multicolumn{1}{c}{\cellcolor[HTML]{ECF4FF}Task-prefix} \\ \hline
\midrule
Perpetrator Mention                & detect perpetrator                       \\ \hline
Perpetrator Type                   & extract perpetrator type                 \\ \hline
Violation Type                     & extract violation type                   \\ \hline
Victim Name                        & extract victims                          \\ \hline
Victim Sex Type                    & extract victim sex                       \\ \hline
Victim Age Group                   & extract victim age                       \\ \hline
Victim Population Type             & extract victim population type           \\ \hline
Victim Type                        & extract victim type                      \\ \hline
Country                            & extract violation country                \\ \hline
Region                             & extract violation region                 \\ \hline
City                               & extract violation city                   \\ \hline
Year                               & extract violation year                   \\ \hline
Month                              & extract violation month                  \\ \hline
Day                                & extract violation day                   \\ \hline
\end{tabular}%
}
\caption{Task Prefix for each event attribute.}
\label{tab:task_prefix}
\end{table}

\section{Long Document Solutions}
\label{appx:long_doc}

Besides the two solutions we evaluate in the paper (Truncation and Knowledge Fusion), there are two other possible solutions that we describe here. First, some work proposes splitting a long document into shorter sequences, then using a transformer to generate sequence representations for each of them \cite{grail2021globalizing}. Then those sequence representations are fed into another network to generate the final document representation. But in this case, a large number of training examples is required to learn the parameters of the network layers which generate the document representation. There is also the long transformer (Longformer \cite{beltagy2020longformer}) approach proposed to handle long documents. But in contrast to the T5 model which is pretrained on many pretraining tasks, it is difficult to reframe all of the subtasks as a unified Sequence-to-Sequence task based on those long transformers. In comparison, the approaches we proposed are all post-processing steps that are less expensive than the aforementioned methods.

\section{Model Performances on the Development Set}
\label{appendix:dev}

\begin{table*}[!ht]
\centering
\resizebox{\textwidth}{!}{%
\begin{tabular}{llcccc}
\hline
\rowcolor[HTML]{ECF4FF}
\cellcolor[HTML]{ECF4FF}Event Attribute                                          & \cellcolor[HTML]{ECF4FF}Metric                & \multicolumn{1}{l}{\cellcolor[HTML]{ECF4FF}DyGIE++} & \cellcolor[HTML]{ECF4FF}T5 w/ Truncation & \cellcolor[HTML]{ECF4FF}T5 w/ \cellcolor[HTML]{ECF4FF}Knowledge Fusion & Hybrid \\ \hline
\midrule
\multicolumn{1}{c}{\multirow{3}{*}{Perpetrator Mention}} & Precision &   \textbf{97.30} &  95.88  &             95.96           &   95.96     \\
\multicolumn{1}{c}{} & Recall                & 37.89  & 97.89 & \textbf{100.00} & 100.00 \\
\multicolumn{1}{c}{} & F1                    &  54.55 & 96.88  & \textbf{97.94} & 97.94 \\ \hline
Perpetrator Type     & Accuracy              &  - & \textbf{68.00} & \textbf{68.00} & 68.00 \\ \hline
\multirow{6}{*}{Victim Name}                             & Exact Match Precision &                10.37                   &           \textbf{85.07}       &              65.08          &     65.08   \\
                     & Exact Match Recall    & 7.14  & 29.08 & \textbf{41.84} & 41.84 \\
                     & Exact Match F1        &  8.46 & 43.35 & \textbf{50.93} & 50.93 \\
                     & Fuzzy Match Precision & 19.26  & \textbf{85.07} & 73.02 & 73.02 \\
                     & Fuzzy Match Recall    & 13.27  & 29.08 & \textbf{46.94} & 46.94 \\
                     & Fuzzy Match F1        &  15.71 & 43.35 & \textbf{57.14} & 57.14 \\ \hline
Victim Type          & Accuracy              & - & \textbf{81.25} & 66.12 & 81.25 \\ \hline
Victim Sex Type      & Accuracy              & - & \textbf{85.42} & 80.87 & 85.42 \\ \hline
Victim Age Group     & Accuracy              & - & 97.92 & \textbf{98.36} & 98.36 \\ \hline
\multirow{3}{*}{Violation Type}                          & Precision             & -                                 &              \textbf{64.14}    &             57.09           &     64.14   \\
                     & Recall                & - & 68.65 & \textbf{76.22} & 68.65 \\
                     & F1                    & - & \textbf{66.32} & 65.28 & 66.32 \\ \hline
Country              & Accuracy  & - & \textbf{62.00} & 59.00 & 62.00 \\ \hline
Region               & Accuracy  & - & \textbf{9.00} & 2.00 & 9.00 \\ \hline
City                 & Accuracy  & - & \textbf{20.00} & 16.00 & 20.00 \\ \hline
Year                 & Accuracy  & - & \textbf{52.00} & 46.00 & 52.00 \\ \hline
Month                & Accuracy  & - & \textbf{32.0}0 & \textbf{32.00} & 32.00 \\ \hline
Day                 & Accuracy  & - & \textbf{18.00} & 10.00 & 18.00 \\ \hline
\end{tabular}%
}
\caption{Overall performance of the baseline models on \dataset dev set (\%). All experiments are based on a single run with a preset random seed.}
\label{tab:results_dev}
\end{table*}
Table~\ref{tab:results_dev} shows the performance of the models on the dev set of \dataset. The best baseline model (T5 Hybrid) is chosen based on the model performance on the dev set.

\end{document}